\documentclass{article}

    \PassOptionsToPackage{numbers, compress}{natbib}

    \usepackage[preprint]{neurips_2024}

\usepackage[utf8]{inputenc} 
\usepackage[T1]{fontenc}    
\usepackage{hyperref}       
\usepackage{url}            
\usepackage{booktabs}       
\usepackage{amsfonts}       
\usepackage{nicefrac}       
\usepackage{microtype}      
\usepackage{xcolor}         

\usepackage{algorithm}
\usepackage{algorithmic}
\usepackage{bbm}
\usepackage{bm}

\usepackage{wrapfig}

\usepackage{times}
\usepackage{graphicx} 
\usepackage{subfigure} 

\usepackage{amsmath}
\usepackage{amssymb}

\usepackage{multirow}

\usepackage{enumitem}
\hypersetup{
    colorlinks=true,
    linkcolor=blue,
    urlcolor=blue,
    hidelinks
}

\usepackage{color, colortbl}
\definecolor{Gray}{gray}{0.9}
\definecolor{demphcolor}{RGB}{144,144,144}
\definecolor{airforceblue}{rgb}{0.36, 0.54, 0.66}
\newcommand{\demph}[1]{\textcolor{demphcolor}{#1}}

\usepackage{subcaption}

\title{Text-only Synthesis for Image Captioning}

\author{%
  Qing Zhou$^\dagger$~~~ Junlin Huang$^\dagger$~~~ Qiang Li~~~ Junyu Gao$^*$~~~ Qi Wang$^*$\\
  School of Artifcial Intelligence, Optics and Electronics (iOPEN) \\
  Northwestern Polytechnical University, Xi'an, China \\
  {\tt\small\{chautsing, lin.pyw, liqmges, gjy3035, crabwq\}@gmail.com} \\
}

\begin{document}

\maketitle

\begin{abstract}
    From paired image-text training to text-only training for image captioning, the pursuit of relaxing the requirements for high-cost and large-scale annotation of good quality data remains consistent. In this paper, we propose \textbf{T}ext-\textbf{o}nly Synthesis for Image \textbf{Ca}ptioning (ToCa), which further advances this relaxation with fewer human labor and less computing time. Specifically, we deconstruct caption text into structures and lexical words, which serve as the fundamental components of the caption. By combining different structures and lexical words as inputs to the large language model, massive captions that contain various patterns of lexical words are generated. This method not only approaches the target domain but also surpasses it by generating new captions, thereby enhancing the zero-shot generalization ability of the model.
    Considering the different levels of data access in the real world, we define three synthesis scenarios: cross-domain synthesis, in-domain synthesis, and data-efficient synthesis.
    Experiments in these scenarios demonstrate the generalizability, transferability and practicability of ToCa with a nearly 5 CIDEr improvement for zero-shot cross-domain captioning and a maximum increase of over 20 CIDEr for data-efficient captioning.

    \end{abstract}

\renewcommand{\thefootnote}{\fnsymbol{footnote}}
\footnotetext[0]{$^\dagger$Authors contributed equally.~~~ $^*$Corresponding author.}
\renewcommand{\thefootnote}{\arabic{footnote}}
\section{Introduction}\label{sec:intro}
Image captioning aims to describe the entities in an image and their interactions using natural language, i.e., what the entities are and how they interact with each other.
Based on extensively annotated paired image-text data, deep learning models \cite{9965949HW, 9308980HW, li2020oscar,BUTD} have demonstrated remarkable performance, fostering applications in the fields of image retrieval and visually impaired assistance \cite{gurari2020captioning}, among others.
However, the annotation of paired image-text data on a large scale presents significant challenges in terms of cost, labor, and time, impeding the widespread application and generalization of models across various scenarios.

To address these challenges, on the one hand, text-only training methods \cite{MAGIC, CapDec, DECAP,ZeroCap,ViECap} leverage the shared semantic representation of CLIP \cite{CLIP} to train zero-shot captioning models using only text data, significantly reducing the difficulty of constructing training data. On the other hand, the paired image-text synthesis methods~\cite{SynTIC,ICSD} utilize diffusion models \cite{StableDiffusion} and large language models (LLMs) \cite{GPT2,GPT3} to generate a large volume of image-text data for training at a remarkably low human cost. However, challenges persist, and the former still requires extensive task-specific text data through manual efforts, while the latter necessitates substantial computational resources to simultaneously generate image-text pairs.

In this paper, we introduce the text-only synthesis task for captioning which aims to further mitigate the need for manually curated datasets and advance the field from paired image-text synthesis to text-only synthesis.
Synthesizing text is simple, but crafting desired text requires skills.
In order to generate desired text in a controlled manner, we propose \textbf{T}ext-\textbf{o}nly Synthesis for Image \textbf{Ca}ptioning (ToCa). Firstly, the captions are deconstructed into meaningful lexical word pairs (representing entities) and structure templates (representing their interactions). Subsequently, LLM is used to generate captions that encompass various interactive behaviors involving these entities. By approaching the target domain, ToCa simultaneously produce novel captions that enhance the model's ability to generalize beyond its original target domain.

ToCa exhibits several distinctive features that contribute to its utility and performance in text synthesis: 1) \textit{Data efficiency}, ToCa can require only a small corpus (even 56 captions) to synthesize a large volume of text data, significantly enhancing model generalization capabilities.
2) \textit{Open accessibility}, the LLM used is open-source and can operate on most graphics cards. This accessibility not only safeguards data privacy but also supports broader usage. Both the code and data will also be openly accessible.
3) \textit{Flexibility}, ToCa allows for the flexible synthesis of any desired text data using different lexical word pairs and structure templates. 4) \textit{Reduced computational steps}, compared to paired image-text synthesis methods, ToCa eliminates the step of image generation (a time-consuming diffusion model generation process), requiring only a 7B LLM to produce text data that meets requirements.

To validate and showcase the formidable synthesis capabilities of ToCa, we conduct a series of extensive experiments based on the accessibility of the data in the real-world. These experiments are categorized as follows: In-domain synthesis, both the data from the corpus and the target dataset are accessible; Cross-domain synthesis, only the data from the corpus is accessible, while the target dataset remained inaccessible; Data-efficient synthesis, limited access to the corpus and the target dataset.
Our main contributions are the following:
\begin{itemize}[itemsep=3pt, topsep=0pt, partopsep=0pt, parsep=0pt]
    \item We propose the text-only synthesis task for the purpose of captioning, aiming to explore methods that require fewer human resources and less computational steps to generate training data for captioning models.
    \item We present ToCa, an algorithm for text synthesis that is flexible, efficient, and completely open-source. It is designed based on the  linguistic composition of caption and greatly improves the model's zero-shot generalization performance.
    \item We establish three common application challenges in real-world, and through these challenges, we validate and demonstrate the impressive synthesis capabilities of ToCa. To the best of our knowledge, ToCa is the first purely synthetic text method for image captioning.
\end{itemize}

\section{Related Works}

\noindent\textbf{Paired image-text training.}
Traditional captioning methods \cite{ClipCap,9857436, ShowAttendTell, alignmentcaption, AoA, BUTD, MMT, CIDEr, SCA-CNN} typically require training on image-text aligned data. During this period of research, the prevalent approach involves employing an encoder-decoder architecture and making subsequent improvements upon it. This involves a transition from CNN-based~\cite{ResNet} visual encoders to ViT-based~\cite{ViT} ones, as well as a shift from LSTM-based~\cite{LSTM} language decoders to Transformers. Various mechanisms~\cite{ShowAttendTell, AoA, SCA-CNN} are explored to enhance the alignment of visual and linguistic features. However, the application of these models is limited by the high cost and scalability challenges associated with manual annotation of paired image-text data. Furthermore, there is a lack of investigation into the zero-shot generalization capabilities of these methods.

\noindent\textbf{Text-only training.}
Regarding the challenge of collecting paired image-text data and the poor generalization of zero-shot image captioning, the text-only training for zero-shot image captioning method has garnered extensive attention and research~\cite{DECAP,ViECap,ZeroCap, CapDec, MAGIC}. The crux of these methods lies in harnessing the formidable image-text coherence representation capability of CLIP, wherein only the text features encoded by CLIP are utilized as input for subsequent training. During the testing phase, CLIP is employed to encode the images into a coherence space, replacing the text features as input for language decoding. Leveraging the generalization capability of CLIP, these methods also demonstrate robust zero-shot capabilities. However, these approaches still rely on a substantial amount of manually annotated text data, and their generalization is constrained by the limitations of such human-generated data. In this paper, we propose a method for synthesizing text data, enabling AI-generated content (AIGC) for text-only training and enhancing zero-shot generalization abilities.

\noindent\textbf{Paired image-text creation.}
Another solution to address the high cost of paired image-text annotation is to leverage automated methods to create image-text data. This can be further divided into two approaches: collecting existing data through web scraping~\cite{CC12M, SimVLM} and synthesizing data~\cite{SynTIC,ICSD,luo2024scalable} using generative models (such as diffusion models and LLM). Feng \emph{et al.}~\cite{SS1M} collected 1 million image-text pairs from the web based on the 80 object categories in COCO dataset \cite{MSCOCO}. ICSD \cite{ICSD} used Stable Diffusion~\cite{StableDiffusion} and ChatGPT3.5 API to synthesize paired image-text data. However, the former approach of web scraping data raises copyright concerns, and in specific tasks, there may not be enough relevant data accessible on the web. The latter approach raises data privacy issues, as ChatGPT3.5 is not open source and is a commercial product. Using data synthesized by ChatGPT3.5 also carries potential commercial risks. Moreover, it is important to consider that each additional step introduces additional risks and time. The effectiveness of Stable Diffusion and LLM affects both the quality and time required for synthesizing the data.

\noindent\textbf{Large language models.}
Following the impressive demonstration of language proficiency by OpenAI's ChatGPT3.5, the open-sourcing of LLM has garnered significant attention and research~\cite{bai2023qwen,cai2024internlm2,abdin2024phi,ai2024yi,bellagente2024stable}. Apart from the pioneering open-source model LLama~\cite{touvron2023llama, llama2}, Mistral has also released their model with high performance and efficient inference~\cite{jiang2023mistral}.
The technique of model quantization~\cite{zhao2023atom,dettmers2208llm} has further enhanced the usability of LLM.
These developments have led us to choose the quantified Mistral and LLama-2 as the LLM for our research in this paper, aiming to further enhance the usability of our methods.

\section{Method}
\subsection{Problem Definition}
For text-only synthesis task, given the corpus $\mathcal{S}$, domain prior information $\mathcal{X}$, where $\mathcal{X}$ could be entities or caption datasets, and the target domain dataset $\mathcal{T}$, our objective is to synthesize a dataset $\mathcal{D}$ that closely approximates or even encompasses the distribution of $\mathcal{T}$ by constructing a synthetic generation strategy $\mathcal{F}$. Formalized as follows:
\begin{equation}
    \begin{aligned}
        &  \min (\mathcal{L}(\mathcal{D}, \mathcal{T}) - \mathcal{L}(\mathcal{D \setminus T}, \mathcal{T})), \\
        & s.t. \quad \mathcal{D} = \mathcal{F}(\mathcal{S}, \mathcal{X}),
    \end{aligned}
    \label{eq:objective}
\end{equation}
where $\mathcal{L}$ represents a measure of the distance between the centers of two data distributions. The term $\min (\mathcal{L}(\mathcal{D}, \mathcal{T}))$ ensures that the generated data $\mathcal{D}$ is close to $\mathcal{T}$, thereby guaranteeing domain-specific performance. Additionally, the term $\min - \mathcal{L}(\mathcal{D \setminus T}, \mathcal{T})$ aims to incorporate information in the generated data $\mathcal{D}$ that is not present in $\mathcal{T}$, thereby enhancing domain-agnostic generalization.
Finding a suitable $\mathcal{L}$ is indeed a challenging task, but it is not the focus of this research.
Instead, we evaluate the synthesis performance by examining the statistical measures between $\mathcal{D}$ and $\mathcal{T}$ (see Appendix~\ref{sec:distance}), as well as the caption performance of the final model (see Sec.~\ref{sec:experiment}).
These evaluations provide insights into the effectiveness of the synthesized data.

Taking into account the challenges arising from variations in data accessibility in real-world, we classify text synthesis into three distinct types based on the availability of $\mathcal{S}$ and $\mathcal{X}$.
In-domain synthesis: in this scenario, we have access to both $\mathcal{S}$ and the target domain data $\mathcal{X=T}$. The objective is to synthesize $\mathcal{D}$ with a focus on generating novel data.
Cross-domain data synthesis: in this case, we have access to $\mathcal{S}$ and a limited set of prior entity relationships $\mathcal{X=R}$. The goal is to synthesize $\mathcal{D}$ by incorporating more domain-specific data.
Data-efficient synthesis: in this situation, we face limitations in terms of the availability of $\mathcal{S}$ and prior information $\mathcal{X}$. The objective is to synthesize $\mathcal{D}$ by efficiently generating a large-scale dataset.

\subsection{Text Synthesis}
Lexical word carry the majority of the semantic information in a sentence, while structure helps to express the relationships between lexical word and construct coherent sentences. In the context of captions, the essence lies in describing the entities and their interactions within an image. Therefore, our core insight is to deconstruct a caption text into two fundamental linguistic elements: lexical pairs and structure templates. By recombining these elements and inputting them into a knowledge-rich LLM, we can achieve the expansion of in-domain text data and the creation of out-of-domain text. Specifically, the text synthesis process involves three steps: structure template construction, lexical pair extraction, and LLM text synthesis.

\noindent\textbf{Structure template construction.}
Here we define that a structure template $G_s$ refers to the basic structure of a sentence $s$, including most function words and part-of-speech (POS) of lexical words, such as \textit{"[N] [VBG] [N] on [N] ."}. For a word $token$, we can write the template function accordingly:
\begin{align}
    g(token) & =
    \begin{cases}
        token & \text{if}\ q_{token} \in \mathcal{Q}_f, \\
        [q_{token}] & \text{if}\ q_{token} \in \mathcal{Q}_c, \\
        None & \text{if else},
    \end{cases}
\end{align}
where $q_{token}$ is the POS of the token, $\mathcal{Q}_c$ is the set of lexical words, and $\mathcal{Q}_f$ is the set of function words. For a sentence $s$, we can obtain its structure template $G_s$:
\begin{align}
    G_s & =  [ g(token_s^0), g(token_s^1), \cdots, g(token_s^n) ],
\end{align}
where $n$ is the token length of sentence $s$.

We employ the use of NLTK\footnote{https://www.nltk.org/} to identify the POSs of words. Taking into consideration the similarities and differences between POSs, lexical words and function words are categorized as  Table~\ref{tab:pos}.

\begin{wraptable}{r}{0.5\textwidth}
    \vspace{-0pt}
    \begin{scriptsize}
        \small
        \caption{The POS sets of lexical words and function words, where $*\in\{B, BD, BG,$ $BN, BP, BZ\}$}
    \centering
    \begin{tabular}{ccc}
    \toprule
    POS Set                & $q_{token}$ & NLTK POS            \\
    \midrule
    \multirow{4}{*}{$\mathcal{Q}_c$} & N   & NN, NNS, NNP, NNPS \\
      & \begin{tabular}[c]{@{}c@{}}V*\end{tabular} & \begin{tabular}[c]{@{}c@{}}V*\end{tabular}                  \\
                       & J   & JJ, JJR, JJS       \\
                       & R   & RB, RBR, RBS       \\
    \midrule
    $\mathcal{Q}_f$ & -    & \begin{tabular}[c]{@{}c@{}}CC, EX, IN, MD, WDT,\\ WP, WP\$, WRB, ``,", ``."\end{tabular} \\
    \bottomrule
    \end{tabular}
    \label{tab:pos}
    \vspace{-5pt}
    \end{scriptsize}
\end{wraptable}

By conducting a statistical analysis of the structure templates of all sentences in the corpus $\mathcal{S}$, we can obtain distributional information about these templates:
\begin{align}
    \mathcal{G} = \{ (G_s, N_{G_s}) | s \in \mathcal{S} \},
\end{align}
where $N_{G_s}$ is the number of occurrences of structure template $G_s$ in $\mathcal{S}$.

\noindent\textbf{Lexical pair extraction.}
The set $R(token^i)$ represents the collection of lexical relationships concerning a specific lexical item $token^i$. It signifies the association between $token^i$  and the set of lexical items that appear after it in the same sentence. By gathering lexical relationships from the corpus, we can more comprehensively restore the entity information and relationships between entities within the sentence.
Specifically, given a sentence $s$ in the corpus $\mathcal{S}$, we can obtain its set of lexical pairs:
\begin{align}
    R_s(token_s^i) = \{ (token_s^j, N_{token_s^i,token_s^j}) | token_s^i, token_s^j \in s, q_{token_s^i}, q_{token_s^j} \in \mathcal{Q}_c, i < j \}
\end{align}
where $N_{token_s^i,token_s^j}$ is the number of occurrences of the lexical pair $(token_s^i, token_s^j)$ in the sentence $s$.
Similarly, by statistically analyzing the lexical pairs of all sentences, we can obtain the distribution information of lexical pairs $\mathcal{R}$:
\begin{align}
    R(token) &= \{ R_s(token) | s \in \mathcal{S}, token \in s , q_{token} \in \mathcal{Q}_c \}, \\
    \mathcal{R} &= \{ (R(token), N_{token}) | s \in \mathcal{S}, token \in s , q_{token} \in \mathcal{Q}_c \},
\end{align}
where $N_{token}$ is the number of occurrences of the lexical word $token$ in the corpus $\mathcal{S}$.

\noindent\textbf{LLM text synthesis.}
Synthesizing text using LLMs is straightforward, yet producing text on a massive scale that meets specific requirements demands sophisticated algorithmic strategies. On one hand, generating a substantial amount of data necessitates a substantial number of prompts. However, we do not wish to manually create a vast number of prompts, as it goes against our original intention. On the other hand, different types and sizes of LLMs possess varying levels of understanding and adherence to prompts. Inadequate prompts result in low-quality desired text. The key lies in having a sufficient quantity of prompts that LLMs can easily follow. By utilizing the structure templates and lexical pairs constructed in the previous context, the ToCa can swiftly generate large-scale prompts through recombination. Another crucial aspect of ToCa is guiding the LLM to complete sentences based on sentence templates, rather than merely constructing them.

Specifically, the process begins by randomly sampling $G_s=[g_s^0,g_s^1,\cdots,g_s^n]$ from the structure template $\mathcal{G}$, based on the size of $N_{G_s}$. Next, for the POS of the first lexical word $g_f^0$ in $G$, a random sampling is performed on the lexical words according to the size of $N_{token}$ to obtain $token_{s,g_f^0}^0$ with the category $g_f^0$. Subsequently, $g_s^0$ in $G_s$ is replaced with $token_{s,g_f^0}^0$. Following this, the lexical category $g_f^1$ of $token_{s,g_f^0}^0$ is sampled based on $R(token_{s,g_f^0}^0)$, and then substituted for $g_s^1$ in $G_s$. This entire process can be represented using the following formula:
\begin{align}
    token_{s,g_f^0} &\thicksim \frac{N_{token_{s,g_f^0}^0}}{\sum\limits_{token_{s,g_f^0}^0} N_{token_{s,g_f^i}}}, \\
    token_{s,g_f^1} &\thicksim \frac{N_{token_{s,g_f^0}^0,token_{s,g_f^1}^1}}{\sum\limits_{token_{s,g_f^1}^1} N_{token_{s,g_f^0}^0,token_{s,g_f^1}^1}}.
\end{align}
For the third and subsequent lexical categories $g_f^i$, it is obvious that their probability distribution should be related to all previous tokens.
We define $[token^{i_0},token^{i_1},\cdots,token^{i_k}]$ to represent these lexical tokens appearing in the sentence in the order of $i_0<i_1<\cdots<i_k$. Then:
\begin{align}
    token_{s,g_f^i}^i\thicksim P(token_{s,g_f^i}^i|[token_{s,g_f^0}^0,token_{s,g_f^1}^1,\cdots,token_{s,g_f^{i-1}}^{i-1}]).
    \label{eq:token_i}
\end{align}
When $i$ is large, accurately computing the conditional probability in Eq.~\ref{eq:token_i} becomes arduous and inefficient. Hence, we employ a straightforward yet potent approximation assumption:
given that $token_{s,g_f^i}^i$ is established, the probabilities of prior token occurrences are independent of their sequence and the events of these tokens occurring are mutually independent.
Thus, we can approximate Eq.~\ref{eq:token_i} as follows:
\begin{equation}
    \begin{aligned}
        & P(token_{s,g_f^i}^i|[token_{s,g_f^0}^0,token_{s,g_f^1}^1,\cdots,token_{s,g_f^{i-1}}^{i-1}]) \\
        =& \frac{P([token_{s,g_f^0}^0,token_{s,g_f^1}^1,\cdots,token_{s,g_f^{i-1}}^{i-1}]|token_{s,g_f^i}^i)P(token_{s,g_f^i}^i)}{P([token_{s,g_f^0}^0,token_{s,g_f^1}^1,\cdots,token_{s,g_f^{i-1}}^{i-1}])} \\
        =& \frac{P(token_{s,g_f^i}^i)\prod\limits_{j=0}^{i-1} P(token_{s,g_f^j}^j|token_{s,g_f^i}^i)}{P(token_{s,g_f^0}^0\cap token_{s,g_f^1}^1\cap\cdots\cap token_{s,g_f^{i-1}}^{i-1})} \\
        =& \frac{(\prod\limits_{j=0}^{i-1} N_{token_{s,g_f^j}^{j},token_{s,g_f^i}^i}) /N_{token_{s,g_f^i}^i}^{i-1}}
        {\sum\limits_{\substack{token_{s,g_f^i}^i}} (\prod\limits_{j=0}^{i-1} N_{token_{s,g_f^j}^{j},token_{s,g_f^i}^i}) /N_{token_{s,g_f^i}^i}^{i-1}}.
    \end{aligned}
    \label{eq:approx}
\end{equation}

From Eq.~\ref{eq:approx}, it can be seen that $N_{token_{s,g_f^i}^i}^{i-1}$ suppresses the excessive sampling of high-frequency words, enhancing the diversity of the data.
To finely adjust the diversity of the data, we introduce a hyperparameter $\tau$ to control the degree of influence of $N_{token_{s,g_f^i}^i}^{i-1}$, where a larger $\tau$ leads to less diversity in sentences:
\begin{equation}
    token_{s,g_f^i}^i\thicksim
    \frac{(\prod\limits_{j=0}^{i-1} N_{token_{s,g_f^j}^{j},token_{s,g_f^i}^i}) /N_{token_{s,g_f^i}^i}^{(i-1)/\tau}}
    {\sum\limits_{token_{s,g_f^i}^i} (\prod\limits_{j=0}^{i-1} N_{token_{s,g_f^j}^{j},token_{s,g_f^i}^i}) /N_{token_{s,g_f^i}^i}^{(i-1)/\tau}}
    \label{eq:token_i_approx}
\end{equation}

In some cases, $N_{token_{s,g_f^j}^{j},token_{s,g_f^i}^i}=0$, for which we simply ignore the current $g_f^i$.

Through the above derivation, we can replace each lexical token in $G_s$ according to the approximate probability distribution of Eq.~\ref{eq:token_i_approx}, thereby obtaining a sentence $s$ containing many lexical tokens.
To utilize the generation ability of LLM, we add MASK markers on both sides of all real words in the obtained $s$, thus obtaining the final sentence template $s'$, such as \textit{"[ ] horse [ ] sitting [ ] field [ ] on [ ] background [ ] ."}.
Then, we can input $s'$ into LLM, requiring LLM to complete the gaps.

Now, from the output of the LLM, we can obtain a complete sentence $s''$ that closely aligns in both style and content with our desired sentence.
If $s''$ does not contain all $token_{s,g_f^i}^i$, it is likely that $s''$ contains some conflicting information.
In this case, we can simply filter out $s''$, which is equivalent to indirectly utilizing the filtering ability of LLM.
By repeating the above process, we can obtain the synthetic dataset $\mathcal{D}={s''_0,s''_1,\cdots,s''_m}$, where $m$ is the preset size of the synthetic dataset.
The upper bound of $m$ is given by:
\begin{align}
    m \leq \sum_{i=1}^{|\mathcal{G}|} {\prod\limits_{j=1}^{n_i} k_{i,j}},
    \label{eq:m_upper_bound}
\end{align}
where $n_{i}$ represents the number of lexical items in the i-th template.
$k_{i,j}$ represents the size of optional lexical items for the j-th $token \in \mathcal{Q}_c$ in the i-th template

\section{Experiment}\label{sec:experiment}
ToCa generates text data in three scenarios: in-domain synthesis, cross-domain synthesis, and data-efficient synthesis. Then assess the performance of models trained on synthesized data in zero-shot image captioning tasks. This evaluation demonstrates the algorithm's generalizability, transferability, and practicability in various contexts.

\noindent\textbf{Implementation details.}
We employ the text-only training method ViECap~\cite{ViECap} as the foundational caption model, keeping the training configuration aligned with it. Regarding the LLM used for synthetic data generation, the 8-bit quantized mistral-7B on huggingface\footnote{https://huggingface.co/TheBloke/Mistral-7B-Instruct-v0.2-GGUF} is used by default. The model undergoes an initial training of 15 epochs on the synthetic data, followed by fine-tuning on the accessible set $\mathcal{X=T}$ (i.e. $\mathcal{D+T}$) and without any fine-tuning on $\mathcal{X=R}$ (i.e. $\mathcal{D}$).

\noindent\textbf{Datasets and metrics.}
Experiments are conducted using the widely recognized image caption benchmarks, COCO~\cite{MSCOCO1}, NoCaps~\cite{NoCaps}, and Flikcr30k~\cite{Flickr30k}, to assess the performance enhancement brought about by the generated data in in-domain captioning and cross-domain captioning tasks. By default, we utilize COCO as the corpus.
COCO test set is used for in-domain caption evaluation. Flikcr30k and NoCaps Val are used for cross-domain caption evaluation.
The dataset partitioning remains consistent with previous works~\cite{ViECap,CapDec} to ensure fairness. The metrics employed are the commonly used BLEU@n (B@n)~\cite{BLEU}, METEOR (M)~\cite{METEOR}, CIDEr (C)~\cite{CIDEr}, and SPICE (S)~\cite{SPICE}.

\subsection{In-domain synthesis}
In this section, we simulate and process real-world scenarios where established or historical data are already created, i.e. $\mathcal{S}=\mathcal{X}=\mathcal{T}=$ COCO training set. In such cases, ToCa performs in-domain synthesis, with the objective of generating novel caption texts to enhance the model's in-domain captioning performance and cross-domain caption generalization performance.

\noindent\textbf{Generalizability.}
As depicted in the Table~\ref{tab:in1}, the ViECap model, trained with data augmentation through ToCa, exhibits remarkable superiority in both in-domain captioning and cross-domain captioning when compared to other text-only training methods. Notably, it is worth highlighting that ToCa provides ViECap with a gain of nearly 5 points in the CIDEr metric for cross-domain captioning. Additionally, in the NoCaps out-of-domain setting, ToCa achieves a significant advancement over the paired image-text training method, surpassing SmallCap by 5.5 points.
Furthermore, it is noteworthy that even though we employ the COCO training set for the extraction of lexical pairs and the construction of structural templates to synthesize text, there is an improvement of 2.1 points in the CIDEr score on the COCO test set. These enhancements demonstrate the effectiveness of ToCa in utilizing the information within the corpus to generate novel, diverse text, thereby infusing the model with out-of-domain knowledge and enhancing its generalization capabilities.
Even compared with the paired image-text synthesis method SynTIC, ToCa has stronger cross-domain generalization.
The small performance gap between ToCa $_\text{llama2-7b}$ and ToCa also indicates that the proposed method has a certain degree of universality to LLM.

\begin{table}
    \vspace{-10pt}
    \caption{\textit{\textbf{Generalizable}} in-domain synthesis. * refers to the use of a memory bank. Except SmallCap reports results on the NoCaps test set, the rest of the methods are on the validation set.}
    \begin{center}
    \small
    \renewcommand\arraystretch{1.15}
    \setlength{\tabcolsep}{1mm}{
    \begin{tabular}{l|cccc|cccc|cccc}
    \toprule
    \multirow{3}{*}{Methods} & \multicolumn{8}{c|}{\textbf{Cross-domain Captioning}} & \multicolumn{4}{c}{\textbf{In-domain Captioning}} \\
    \cmidrule{2-13}
    ~ & \multicolumn{4}{c|}{Flickr30k} & \multicolumn{4}{c|}{NoCaps Val (CIDEr)} & \multicolumn{4}{c}{COCO} \\
    ~ & B@4 & M & C & S & In & Near & Out & Overall & B@4 & M & C & S \\
    \midrule
    \multicolumn{13}{l}{\demph{\textit{Paired image-text training}}} \\
    \demph{ClipCap~\cite{ClipCap}} \demph{$_\text{\rm ArXiv'21}$} & \demph{-} & \demph{-} & \demph{-} & \demph{-} & \demph{84.9} & \demph{66.8} & \demph{49.1} & \demph{65.8} & \demph{33.5} & \demph{27.5} & \demph{113.1} & \demph{21.1} \\
    \demph{I-Tuning$_{\rm Base}$\ ~\cite{ITuning}} \demph{$_\text{\rm ICASSP'23}$} & \demph{-} & \demph{-} & \demph{-} & \demph{-} & \demph{83.9} & \demph{70.3} & \demph{48.1} & \demph{67.8} & \demph{34.8} & \demph{28.3} & \demph{116.7} & \demph{21.8} \\
    \demph{SmallCap*~\cite{SMALLCAP}} \demph{$_\text{\rm CVPR'23}$} & \demph{-} & \demph{-} & \demph{-} & \demph{-} & \demph{83.3} & \demph{77.1} & \demph{65.0} & \demph{75.8} & \demph{37.0} & \demph{27.9} & \demph{119.7} & \demph{21.3} \\

    \multicolumn{13}{l}{\demph{\textit{Paired image-text synthesis}}} \\
    \demph{ICSD~\cite{ICSD}} \demph{$_\text{\rm AAAI'24}$} & \demph{-} & \demph{-} & \demph{-} & \demph{-} & \demph{-} & \demph{-} & \demph{-} & \demph{-} & \demph{29.9} & \demph{25.4} & \demph{96.6} & \demph{-} \\
    \demph{SynTIC~\cite{SynTIC}} \demph{$_\text{\rm AAAI'24}$} & \demph{17.9} & \demph{18.6} & \demph{38.4} & \demph{11.9} & \demph{-} & \demph{-} & \demph{-} & \demph{-} & \demph{29.9} & \demph{25.8} & \demph{101.1} & \demph{19.3} \\

    \multicolumn{13}{l}{\textit{Text-only training}} \\
    DeCap*~\cite{DECAP} \demph{$_\text{\rm ICLR'22}$} & 16.3 & 17.9 & 35.7 & 11.1 & \textbf{65.2} & 47.8 & 25.8 & 45.9 & 24.7 & 25.0 & 91.2 & 18.7 \\
    CapDec~\cite{CapDec} \demph{$_\text{\rm EMNLP'22}$} & 17.3 & 18.6 & 35.7 & - & 60.1 & 50.2 & 28.7 & 45.9 & 26.4 & 25.1 & 91.8 & - \\
    \rowcolor{Gray}
    ViECap~\cite{ViECap} \demph{$_\text{\rm ICCV'23}$} & 17.4 & 18.0 & 38.4 & 11.2 & 61.1 & 64.3 & 65.0 &  66.2 & \textbf{27.2} & 24.8 & 92.9 & 18.2 \\
    \rowcolor{Gray}
    \textbf{+ToCa$_\text{llama2-7b}$} & \underline{17.7}   & \textbf{18.7}   & \underline{42.7}   & \underline{12.3}  & 62.5 & \underline{67.8} & \underline{69.1} & \underline{69.5} &  \textbf{27.2}   & \underline{25.3}   & \underline{94.0}   & \underline{18.8}  \\
    \rowcolor{Gray}
    \textbf{+ToCa} & \textbf{18.2} & \textbf{18.7} & \textbf{43.9} & \textbf{12.6} & \underline{64.6} & \textbf{69.1} & \textbf{70.5} & \textbf{70.9} & \underline{27.1} & \textbf{25.4} & \textbf{95.0} & \textbf{19.0}  \\
    \bottomrule
    \end{tabular}}
    \end{center}
    \label{tab:in1}
    \end{table}

\begin{table}
    \vspace{-18pt}
    \caption{\textit{\textbf{Flexible}} in-domain synthesis}
    \vspace{4pt}
    \centering
    \small
    \begin{minipage}{0.6\textwidth}
        \centering
        \subtable[FlickrStyle10K]{
        \setlength{\tabcolsep}{.9mm}{
        \begin{tabular}{l|cccc|cccc}
        \toprule
        \multirow{2}{*}{Method} & \multicolumn{4}{c|}{\textbf{Romantic}} & \multicolumn{4}{c}{\textbf{Humorous}}\\
        ~ & B@1 & B@3 & M & C & B@1 & B@3 & M & C \\
        \midrule
        StyleNet~\cite{StyleNet} & 13.1 & 1.5 & 4.5 & 7.2 & 13.4 & 0.9 & 4.3 & 11.3 \\
        MemCap~\cite{MemCap} & 21.2 & 4.8 & 8.4 & 22.4 & 19.9 & 4.3 & 7.4 & 19.4 \\
        CapDec~\cite{CapDec} & 21.4 & 5.0 & 9.6 & 26.9 & 24.9 & 6.0 & 10.2 & 34.1 \\
        \rowcolor{Gray}
        ViECap~\cite{ViECap} & 25.7 & 6.5 & 10.4 & 33.6 & 24.3 & 6.5 & 10.4 & 35.0 \\
        \rowcolor{Gray}
        \textbf{+ToCa} & \textbf{27.6} & \textbf{8.5} & \textbf{11.7} & \textbf{41.4} & \textbf{26.7} & \textbf{8.0} & \textbf{11.5} & \textbf{41.5} \\
        \bottomrule
        \end{tabular}}
        \label{tab:flickrstyle10k}
        }
    \end{minipage}\hfill
    \begin{minipage}{0.4\textwidth}
        \centering
        \subtable[MSR-VTT]{
        \setlength{\tabcolsep}{.9mm}{
        \begin{tabular}{l|cccc}
        \toprule
        Method & B@4 & M & C & S \\
        \midrule
        MAGIC~\cite{MAGIC} & 5.5 & 13.3 & 7.4 & - \\
        CLIPRe~\cite{MAGIC} & 10.2 & 18.8 & 19.9 & - \\
        DeCap~\cite{DECAP} & 23.1 & 23.6 & 34.8 & - \\
        CapDec~\cite{CapDec} & 22.2 & 22.5 & 29.4 & \textbf{6.4} \\
        \rowcolor{Gray}
        ViECap~\cite{ViECap} & 23.3 & 21.5 & 29.3 & 5.3 \\
        \rowcolor{Gray}
        \textbf{+ToCa} & \textbf{28.8} & \textbf{24.2} & \textbf{36.5} & 6.3 \\
        \bottomrule
        \end{tabular}
        }
        \label{tab:MSR-VTT}
        }
    \end{minipage}
    \vspace{-20pt}
  \end{table}

\begin{wraptable}{r}{0.4\textwidth}
  \begin{center}
  \small
  \vspace{-0pt}
  \caption{\textit{\textbf{Transferable}} cross-domain \\synthesis.}
  \setlength{\tabcolsep}{1.2mm}{
  \begin{tabular}{l|cccc}
  \toprule
  \multirow{2}{*}{Method} & \multicolumn{4}{c}{\textbf{Flickr30k $\Rightarrow$ COCO}}\\
  ~ & B@4 & M & C & S  \\
  \midrule
  MAGIC~\cite{MAGIC} & 5.2 & 12.5 & 18.3 & 5.7 \\
  DeCap~\cite{DECAP} & 12.1 & 18.0 & 44.4 & 10.9 \\
  CapDec~\cite{CapDec} & 9.2 & 16.3 & 27.3 & - \\
  \rowcolor{Gray}
  ViECap~\cite{ViECap} & 12.6 & 19.3 & 54.2 & 12.5 \\
  \rowcolor{Gray}
  +$\text{ToCa}_{\mathcal{S}}$ & \textbf{14.8} & \textbf{20.4} & \textbf{58.9} & \textbf{13.9} \\
  \midrule
  +$\text{ToCa}_{\mathcal{S+X}}$ & 15.5 & 21.3 & 61.5 & 14.8 \\
  \bottomrule
\end{tabular}}
\label{tab:crossdomain}
\end{center}
\vspace{-10pt}
\end{wraptable}

\noindent\textbf{Flexibility.}
To demonstrate the flexibility and controllability of ToCa in text synthesis, we apply it to the tasks of style captioning (FlickrStyle10k~\cite{StyleNet}) and video captioning (MSR-VTT~\cite{MSRVTT}).
From Table~\ref{tab:flickrstyle10k}, it is evident that ToCa significantly improves the model's performance in both romantic and humorous captioning. This underscores the flexibility and controllability of ToCa in style captioning.
From Table~\ref{tab:MSR-VTT}, it can be observed that ToCa consistently enhances captioning performance even in the context of video captioning tasks. This indicates the versatility of ToCa across different tasks.
We attribute these improvements to the construction of task-independent lexical pairs and structural templates by ToCa. By deconstructing captions from a linguistic perspective, ToCa enables flexible synthesis of any desired text.

\subsection{Cross-domain synthesis}
In this section, we explore the impact of synthesized text on cross-domain captioning when the target dataset $\mathcal{T}$, specifically the COCO training set, is inaccessible, while a certain amount of similar corpus $\mathcal{S}$, such as Flickr30k, is accessible. As shown in the Table~\ref{tab:crossdomain}, ToCa continues to significantly outperform other models, exhibiting a CIDEr score advantage of 4.7. This demonstrates the remarkable cross-domain \textbf{transferability} of ToCa. Additionally, when there is some prior information $\mathcal{X}$ available, such as COCO training set lexical pairs, the cross-domain effectiveness of ToCa can be further enhanced ($\text{ToCa}_{\mathcal{S+X}}$). This indicates that ToCa possesses good synthesis controllability, enabling effective integration of prior information. The manual cost associated with creating such prior information is much lower compared to constructing a complete dataset, as it only requires a few word pairs rather than an entire sentence.

\begin{wraptable}{r}{0.5\textwidth}
    \small
    \vspace{-10pt}
    \caption{\textit{\textbf{Practical}} data-efficient synthesis.}
    \vspace{-4pt}
    \begin{center}
    \setlength{\tabcolsep}{0.9mm}{
    \begin{tabular}{r|c|c|cccc}
    \toprule
    \multirow{2}{*}{Data} & \multirow{2}{*}{Method} &\multicolumn{1}{c|}{\textbf{COCO}}&\multicolumn{4}{c}{\textbf{NoCaps val}} \\
     & & Test & In & Near & Out & Overall \\
    \midrule
    \multirow{3}{*}{0.01\%} & CapDec    & 0.5  &  0.1 &  0.1 &  0.1  & 0.1 \\
    & ViECap &   0.3 &   0.1   &    0.1    &    0.1   &   0.1  \\
    & ToCa  &   \textbf{21.3}   &   \textbf{15.5}  &   \textbf{ 17.9}   &   \textbf{18.0}   &  \textbf{18.6} \\
    \midrule
    \multirow{3}{*}{0.1\%} & CapDec & 24.0 & 13.2 & 11.0 & 6.2 & 10.4 \\
    & ViECap & 32.3 & 20.9 & 27.6 & 34.9 & 30.2 \\
    & ToCa  & \textbf{49.8}  & \textbf{37.5}  & \textbf{37.0}   & \textbf{39.4} & \textbf{39.6} \\
    \midrule
    \multirow{3}{*}{1\%} & CapDec & 55.8 & 29.6 & 20.5 & 9.8 & 18.9 \\
    & ViECap & 63.9 & 34.6 & 39.9 & 39.3 & 40.4\\
    & ToCa & \textbf{80.9}  & \textbf{57.9} & \textbf{58.6}  & \textbf{59.2} & \textbf{60.5}\\
    \midrule
    \multirow{3}{*}{10\%} & CapDec & 83.6 & 47.3 & 39.8 & 19.1& 35.4\\
    & ViECap & 83.4 & 45.9 & 51.8 & 48.7  & 53.3 \\
    & ToCa & \textbf{91.8}  & \textbf{59.2} & \textbf{64.0}  & \textbf{62.7} & \textbf{65.0} \\
    \midrule
    \multirow{3}{*}{100\%} & CapDec & 92.7 & 60.1 & 50.2 & 28.7 & 45.9  \\
    & ViECap & 92.9 & 61.1 & 64.3 & 65.0 & 66.2 \\
    & ToCa & \textbf{95.0} & \textbf{64.6}  & \textbf{69.1} & \textbf{70.5} & \textbf{70.9} \\
    \bottomrule
    \end{tabular}}
    \end{center}
    \label{tab:fewshot}
    \vspace{-18pt}
\end{wraptable}

\subsection{Data-efficient synthesis}
In this section, we discuss text synthesis with low quantities of data.
Following the ViECap setting, we randomly select data from different scales in the COCO training set for text synthesis and training.
As shown in the Table~\ref*{tab:fewshot}, regardless of the scale, ToCa significantly improves the performance of the model in both in-domain captioning and cross-domain captioning, with the maximum improvement being 20 CIDEr score at the 1\% scale. It is surprising that ViECap, based on ToCa's synthesized data augmentation, achieves comparable performance to ViECap with ten times the amount of data at any scale ($>0.1\%$). Even when the data volume is only 0.01\% (\textbf{only 56 texts}), ToCa is still able to synthesize 1,076 caption texts without repetition, greatly enhancing the captioning performance. These improvements are attributed to ToCa's structure and recombination strategy, which provide ample prompts for LLM synthesis of texts, even with small data volumes.

\subsection{Ablation Studies}
\begin{figure}
  \vspace{-18pt}
  \centering
  \begin{subfigure}[COCO test set]{
    \includegraphics[width=0.3\textwidth]{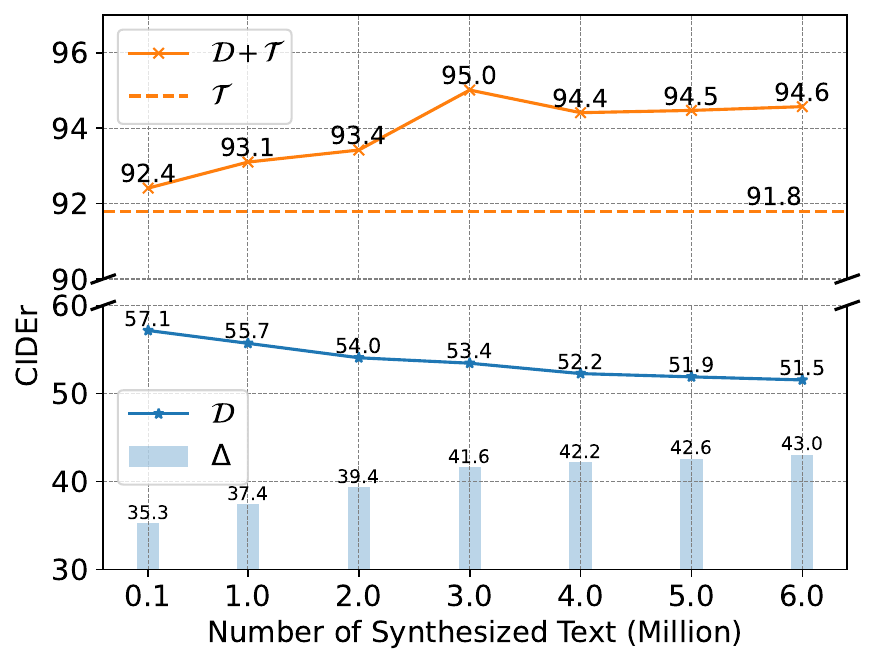}
    \label{fig:num-coco}
    }
  \end{subfigure}
  \hfill
  \begin{subfigure}[NoCaps overall]{
    \includegraphics[width=0.3\textwidth]{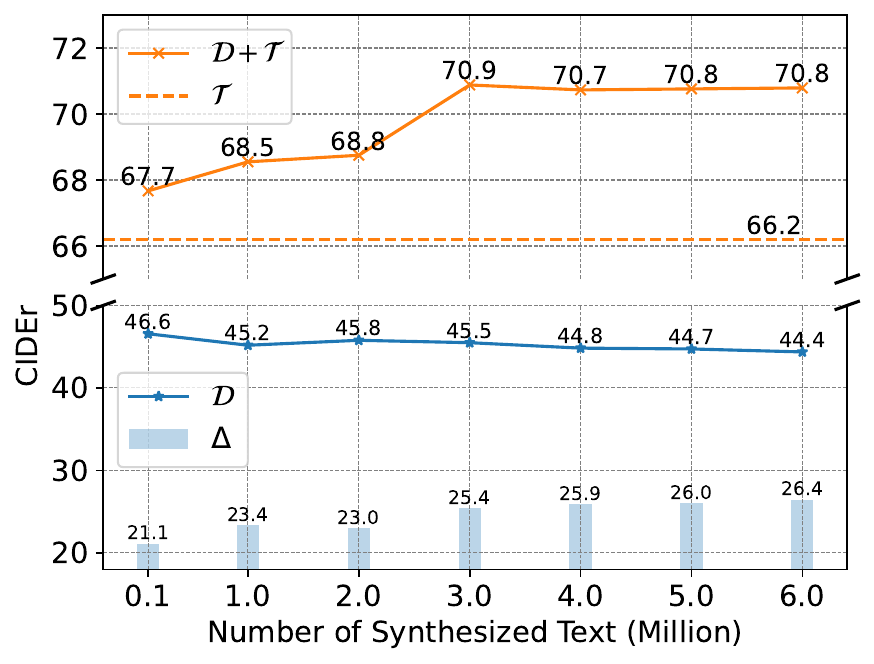}
    \label{fig:num-nocaps}
    }
  \end{subfigure}
  \hfill
  \begin{subfigure}[Different $\tau$]{
    \includegraphics[width=0.3\textwidth, height=0.1397\textheight]{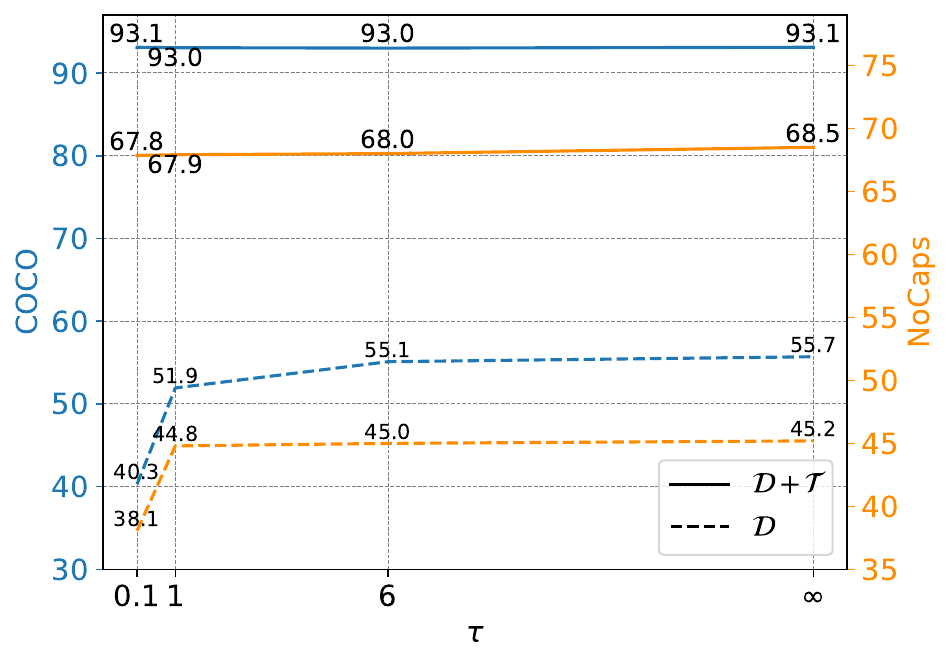}
    \label{fig:tau}
    }
  \end{subfigure}
  \vspace{-8pt}
  \caption{The effect of the number of synthesized text and different $\tau$. $\mathcal{D+T}$ represents training on synthesized data $\mathcal{D}$ and fine-tuning on accessible target data $\mathcal{T}$. $\mathcal{D}$ represents training solely on synthesized data. $\Delta$ denotes the difference in CIDEr scores between $\mathcal{D+T}$ and $\mathcal{D}$.}
  \label{fig:num}
\end{figure}

\noindent\textbf{The effect of synthesis volume.}
We conduct experiments to evaluate the performance of model under different synthetic data volumes, as shown in Figure~\ref{fig:num}. The observation showed a steady improvement in model performance from 0.1M to 3M data under $\mathcal{D+T}$, but it reached a plateau beyond 3M.
We believe that ToCa is still capable of generating diverse novel texts (based on Eq.~\ref{eq:m_upper_bound}), but not relevant to the target dataset, which leads to no performance gain.
The experimental results also support this statement: in Figure~\ref{fig:num-coco}, models trained only on $\mathcal{D}$ show a decreasing CIDEr score on COCO as the data volume increases. This is because after synthesizing strongly correlated and high-frequency texts, further increasing the quantity leads to weakly correlated and low-frequency novel texts (noise). In contrast, Figure~\ref{fig:num-nocaps} shows more stability on NoCaps, as NoCaps is not part of the corpus data, so the correlation between the synthesized text and NoCaps captions remains stable regardless of the scale of synthesis. The higher CIDEr score on COCO compared to NoCaps also indicates that ToCa does indeed approach the target dataset to some extent.

\begin{wraptable}{r}{0.5\textwidth}
    \small
    \vspace{-6pt}
    \caption{The effect of lexical pairs $\mathcal{R}$ and structure templates $\mathcal{G}$.}
    \vspace{-6pt}
    \begin{center}
    \setlength{\tabcolsep}{0.9mm}{
    \begin{tabular}{c|c|c|cccc}
    \toprule
    \multirow{2}{*}{Method} & \multirow{2}{*}{Data} &\multicolumn{1}{c|}{\textbf{COCO}}&\multicolumn{4}{c}{\textbf{NoCaps val}} \\
     & & Test & In & Near & Out & Overall \\
    \midrule
    Baseline & $\mathcal{T}$    & 91.8   & 60.9   & 63.7 &  64.2 &  65.4 \\
    \midrule
    \multirow{2}{*}{$\mathcal{R}$ w/o}  & $\mathcal{D}$ & 19.9 & 13.0 & 16.1 &  14.6 & 15.9 \\
        & $\mathcal{D+T}$ & 92.2 & 61.0 & 64.7 & 65.2 & 66.5 \\
    \midrule
    \multirow{2}{*}{$\mathcal{G}$ w/o}  & $\mathcal{D}$   & 54.3 & 39.3 & 42.7 & 41.7 & 43.2 \\
       & $\mathcal{D+T}$ & 93.0 & 61.5 & 66.4 & 67.2 & 68.3 \\
    \midrule
    \multirow{2}{*}{ToCa}  & $\mathcal{D}$   & 55.7 & 42.7 & 44.1 & 43.7 & 45.2 \\
        & $\mathcal{D+T}$ & \textbf{93.1} & \textbf{61.8}  & \textbf{66.5} & \textbf{68.4} & \textbf{68.5} \\
    \bottomrule
    \end{tabular}}
    \end{center}
    \label{tab:ablation}
    \vspace{-10pt}
    \end{wraptable}

\noindent\textbf{The effect of deconstruction strategy.}
In order to investigate the impact of deconstructing captions into lexical pairs $\mathcal{R}$ and structure templates $\mathcal{G}$ on synthesized text, we conducted experiments on a dataset of 1M, as shown in Table~\ref{tab:ablation}. It can be observed that the removal of $\mathcal{R}$ results in a significant decrease in performance on the dataset $\mathcal{D}$, indicating the crucial importance of lexical pairs as the core meaning of captions for synthesizing text in the target domain. Although the removal of $\mathcal{G}$ does not lead to a significant decrease on the dataset $\mathcal{D+T}$, there is a decline on the dataset $\mathcal{D}$, suggesting that structure templates also play a controlling role in synthesizing text in the target domain.
By considering Eq.~\ref{eq:token_i_approx} in conjunction with Figure~\ref{fig:tau}, it becomes apparent that selecting a smaller value for $\tau$ can result in an excessive diversification of the synthesized data, thereby straying too far from the realm of $\tau$ and introducing noise that ultimately diminishes performance. Although training on $\mathcal{D+T}$ mitigates a substantial portion of these discrepancies, $\tau=\infty$ is chosen by default for more robust performance, as it aligns more closely with our objective function Eq.~\ref{eq:objective}.

\begin{figure}
  \vspace{0pt}
  \includegraphics[width=\textwidth, height=0.37\textheight]{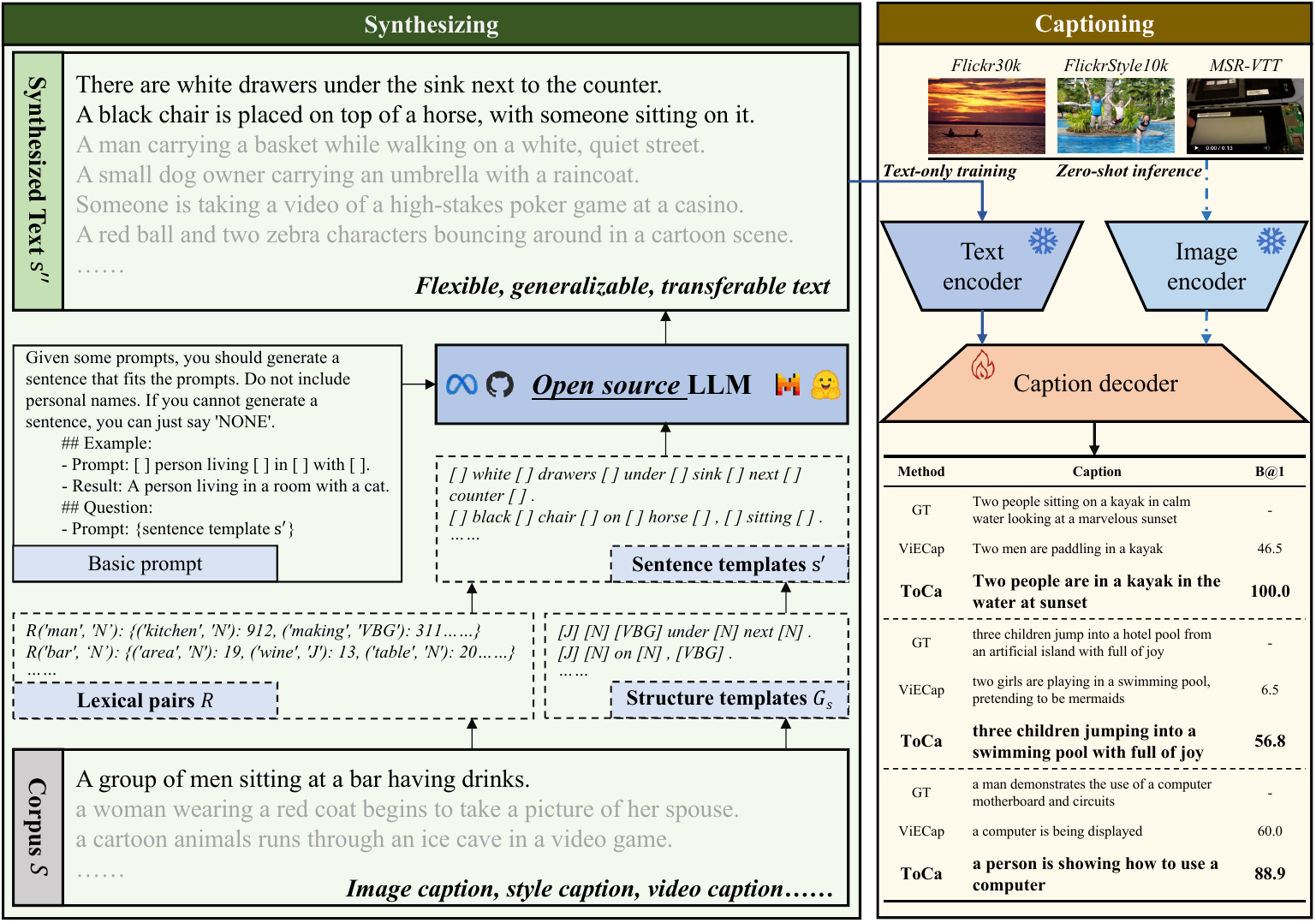}
  \caption{Visualization of the process and results of synthesizing and captioning.}
  \label{fig:overall}
  \vspace{-8pt}
\end{figure}

\subsection{Qualitative visualization}
We visually showcase the qualitative results of the synthesized text and captioning results, as depicted in the Figure~\ref{fig:overall}. On the left side of the Figure~\ref{fig:overall}, the synthesizing results demonstrate the ability of the generated text to effectively incorporate entity information, adapt to different styles and tasks. On the right side of the Figure~\ref{fig:overall}, the captioning results reveal the improved descriptive outcomes achieved through the process of enhanced training. These findings intuitively underscore the potential and advantages of text synthesis in elevating the quality and relevance of caption generation. More synthesis results and analyses can be viewed in the Appendix~\ref{sec:distance} and Appendix~\ref{sec:examples}.
\section{Conclusion}
In this paper, we propose the ToCa, which takes the lead in utilizing the LLM to synthesize caption text. This pioneering approach has yielded conspicuous  performance gains and cost reductions across various real-world challenges and a diverse range of caption tasks. We perceive LLM as a comprehensive repository, encapsulating a multitude of knowledge, while our algorithm constructs a refined cue to retrieve suitable information. Consequently, our aspiration for ToCa extends beyond facilitating the advancement of the visual caption domain; we aim to inspire a broader spectrum of text generation tasks and various types of data synthesis based on synthesized text prompts.
In the future, we will explore and refine ToCa's application in these domains, promoting the achievement of the paradigm of \textit{AIGC in, AIGC out}.

\clearpage

\bibliographystyle{abbrv}
{
	\small
	\bibliography{ToCa}
}

\appendix

\newpage

\section{Limitations}
\label{sec:limitations}
Although we have demonstrated the powerful capabilities of ToCa in text synthesis through various real-world challenges and diverse tasks, limitations in terms of time and workload have prevented us from conducting experiments in additional task scenarios, such as dense captioning and paragraph captioning. Furthermore, a specific metric $\mathcal{L}$ to measure the distance between the centers of two data distributions has not been extensively discussed or explored.
Instead, we indirectly reflect on this measure through the final performance metrics, as well as through direct quantitative statistics and qualitative visualizations.

\section{Impacts}
\label{sec:impacts}
\noindent\textbf{Positive Impacts.}
The scarcity of data samples poses a significant obstacle to the application of artificial intelligence technology in many real-world scenarios. However, ToCa proves to be effective in domains with limited data samples, facilitating the training of robust models and promoting the achievement of the paradigm of \textit{AIGC in, AIGC out}.

\noindent\textbf{Negative Impacts.}
LLMs themselves possess the potential challenges of discrimination, exclusion, and toxicity, where synthetic text may be subject to interference and impact models trained on such data. Additionally, due to the high flexibility of ToCa, malicious input in the form of a corrupted corpus could potentially result in negative text generation. Nevertheless, as the research community delves deeper into addressing biases and related concerns surrounding LLMs, most LLMs exhibit a degree of self-monitoring, enabling them to filter or even cease response to malicious inputs.

\section{Implementation details}
\label{sec:exp-details}
ViECap is used as the base caption model, ensuring that all training and inference settings remained consistent with ViECap. The data splits for COCO and Flickr30k followed the widely adopted Karpathy split~\cite{alignmentcaption}. For evaluating NoCaps, following the convention~\cite{Oscar,ViECap} of using the validation set to assess models trained on COCO. As for FlickrStyle10k, following the processing approach of ViECap~\cite{ViECap} and MemCap~\cite{MemCap}, randomly selecting 6,000 samples for the training set while keeping the remainder for the test set. Regarding MSR-VTT, we adhered to the partitioning method described in~\cite{yu2018joint,luo2022clip4clip} and maintained consistency with the comparative approaches. LLM inference and text generation are performed using the llama-cpp-python\footnote{https://github.com/abetlen/llama-cpp-python} project code.

The Mistral-7B model, quantized to 8 bits, demands approximately 8GB of GPU memory. When executed on an NVIDIA GeForce RTX 3090 GPU, it takes approximately 1 seconds to generate a single sentence. Leveraging the advantages of quantization technology, a 3090 GPU (24GB memory) enables the simultaneous execution of two processes, thereby accelerating the generation process.

\section{Distance between $\mathcal{D}$ and $\mathcal{T}$}
\label{sec:distance}

\begin{figure}
    \vspace{-18pt}
    \centering
    \begin{subfigure}[Flickr30k / COCO = 0.1]{
      \includegraphics[width=0.3\textwidth]{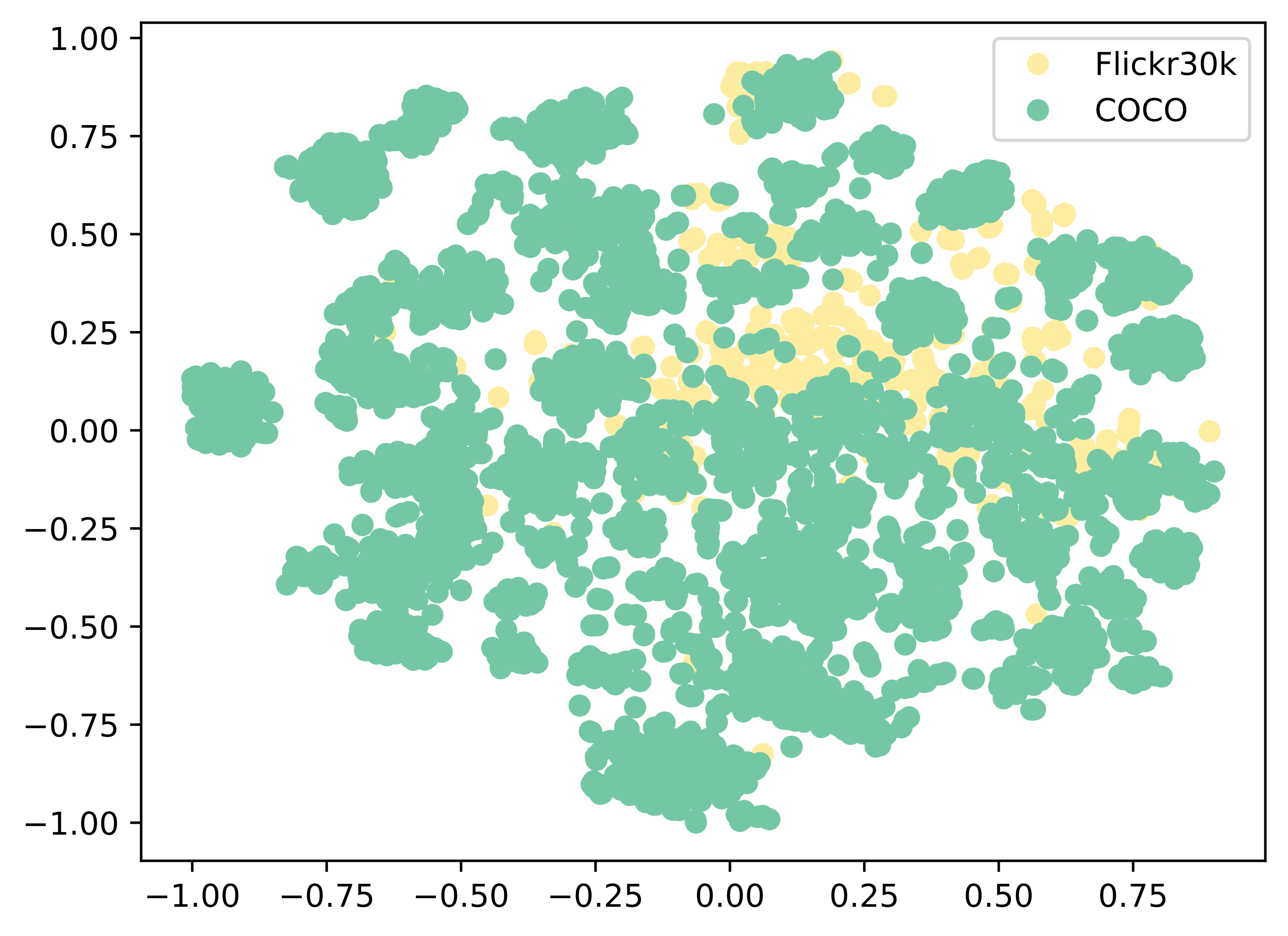}
      \label{fig:coco-flickr30k-0.1x}
      }
    \end{subfigure}
    \hfill
    \begin{subfigure}[Flickr30k / COCO = 1]{
      \includegraphics[width=0.3\textwidth]{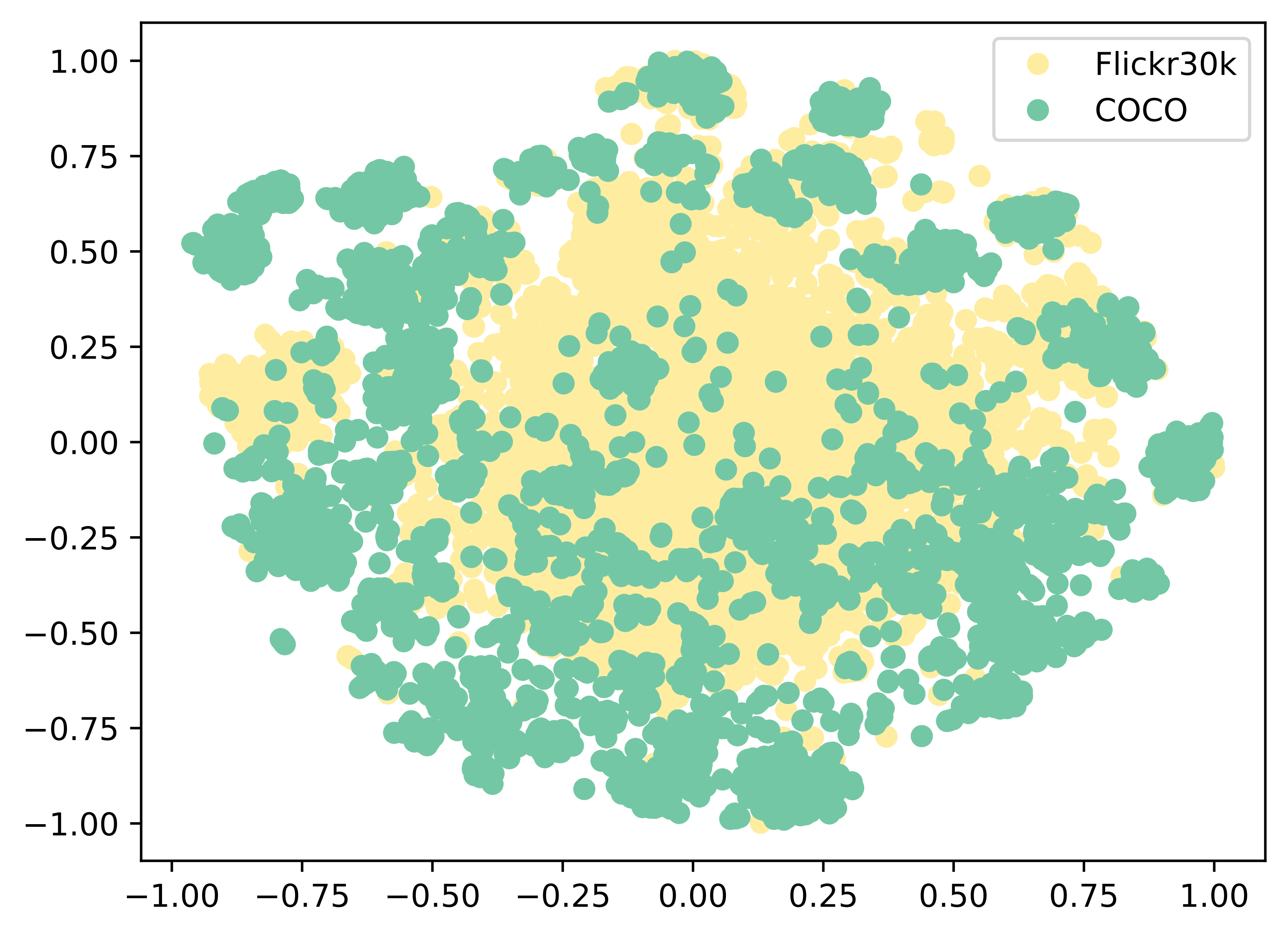}
      \label{fig:coco-flickr30k-1x}
      }
    \end{subfigure}
    \hfill
    \begin{subfigure}[Flickr30k / COCO = 10]{
      \includegraphics[width=0.3\textwidth]{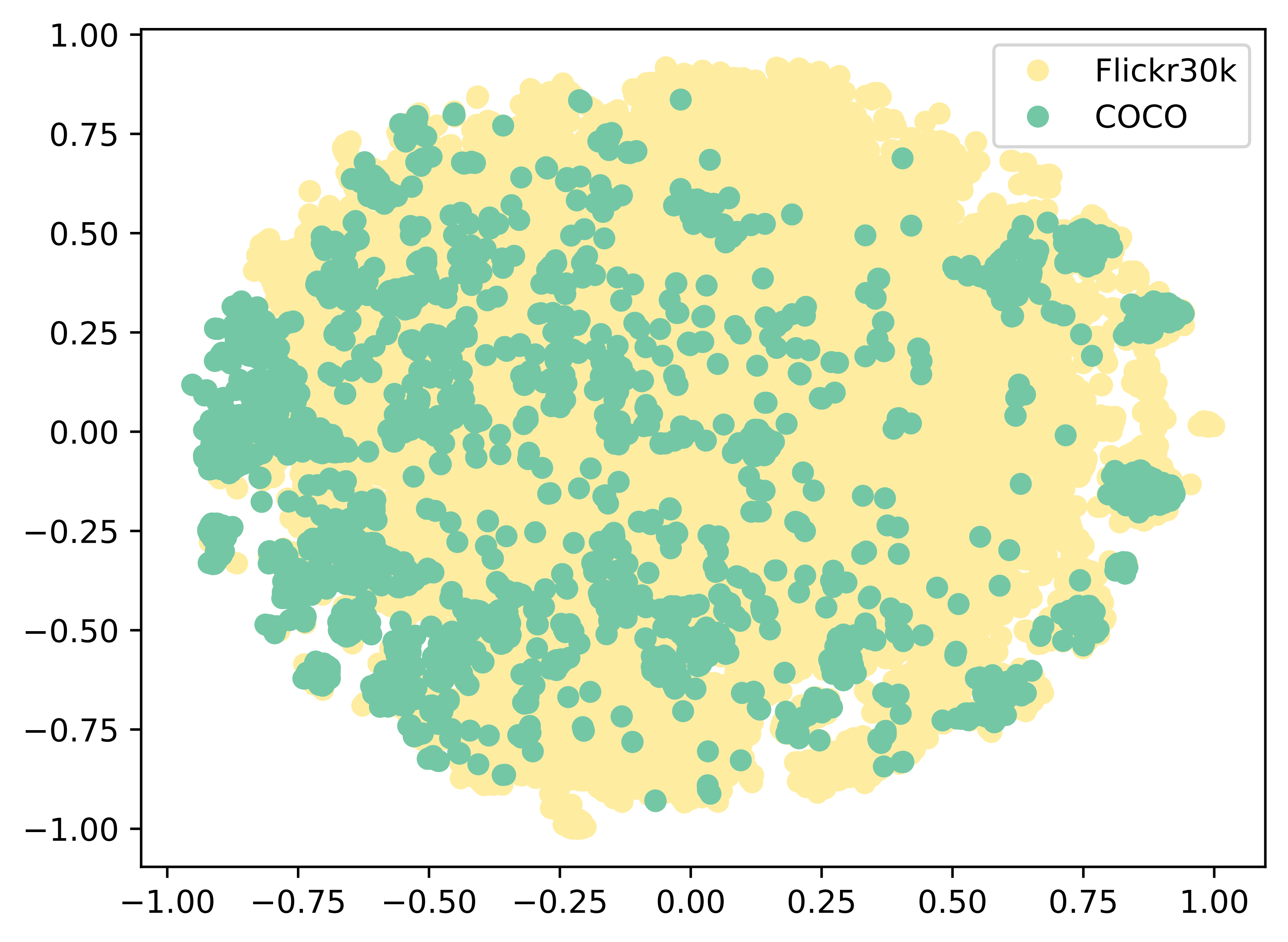}
      \label{fig:coco-flickr30k-10x}
      }
    \end{subfigure}

    \hfill
    \begin{subfigure}[ToCa / COCO = 0.1]{
        \includegraphics[width=0.3\textwidth]{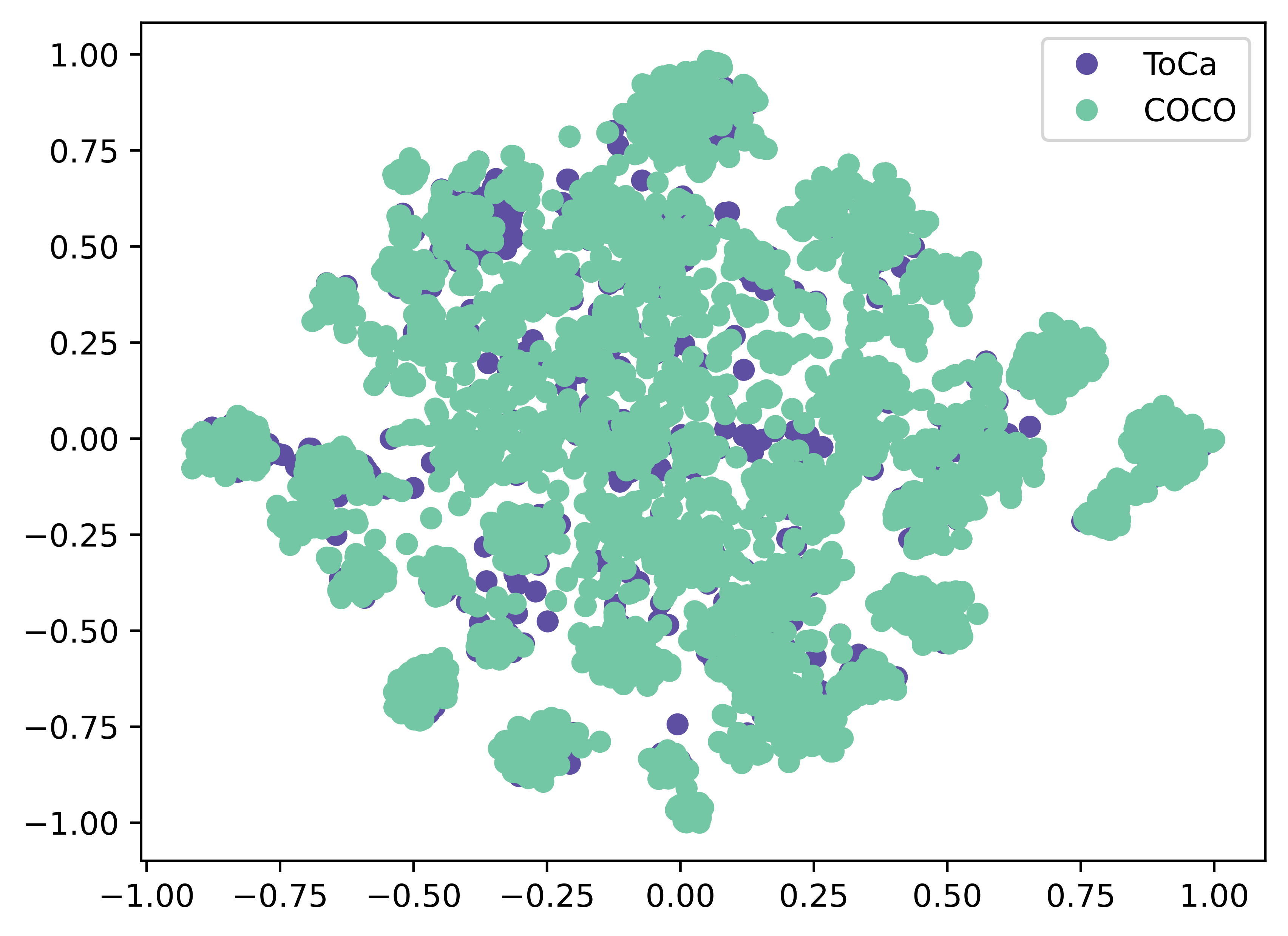}
        \label{fig:coco-toca-0.1x}
        }
      \end{subfigure}
      \hfill
    \begin{subfigure}[ToCa / COCO = 1]{
        \includegraphics[width=0.3\textwidth]{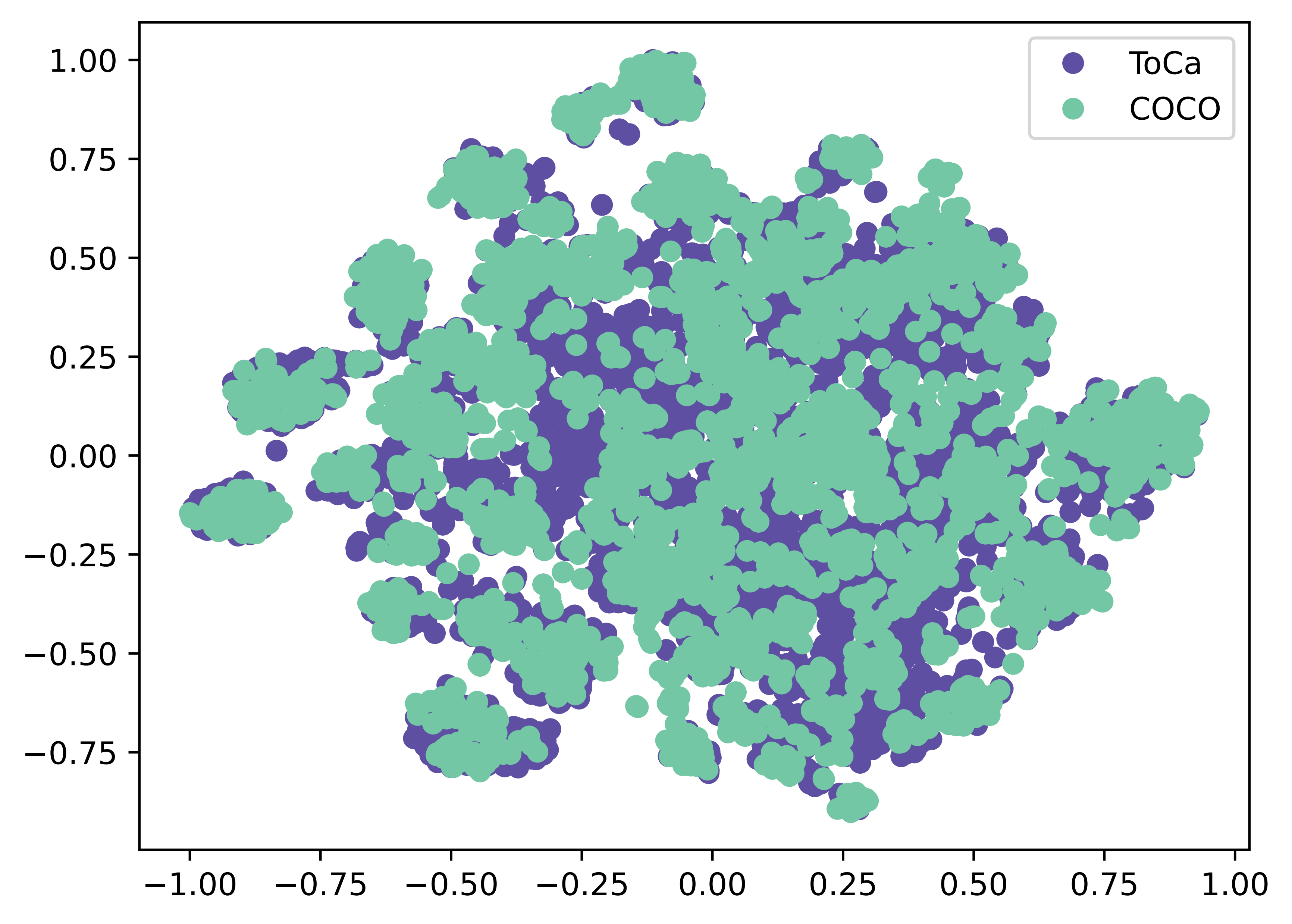}
        \label{fig:coco-toca-1x}
        }
      \end{subfigure}
      \hfill
      \begin{subfigure}[ToCa / COCO = 10]{
        \includegraphics[width=0.3\textwidth]{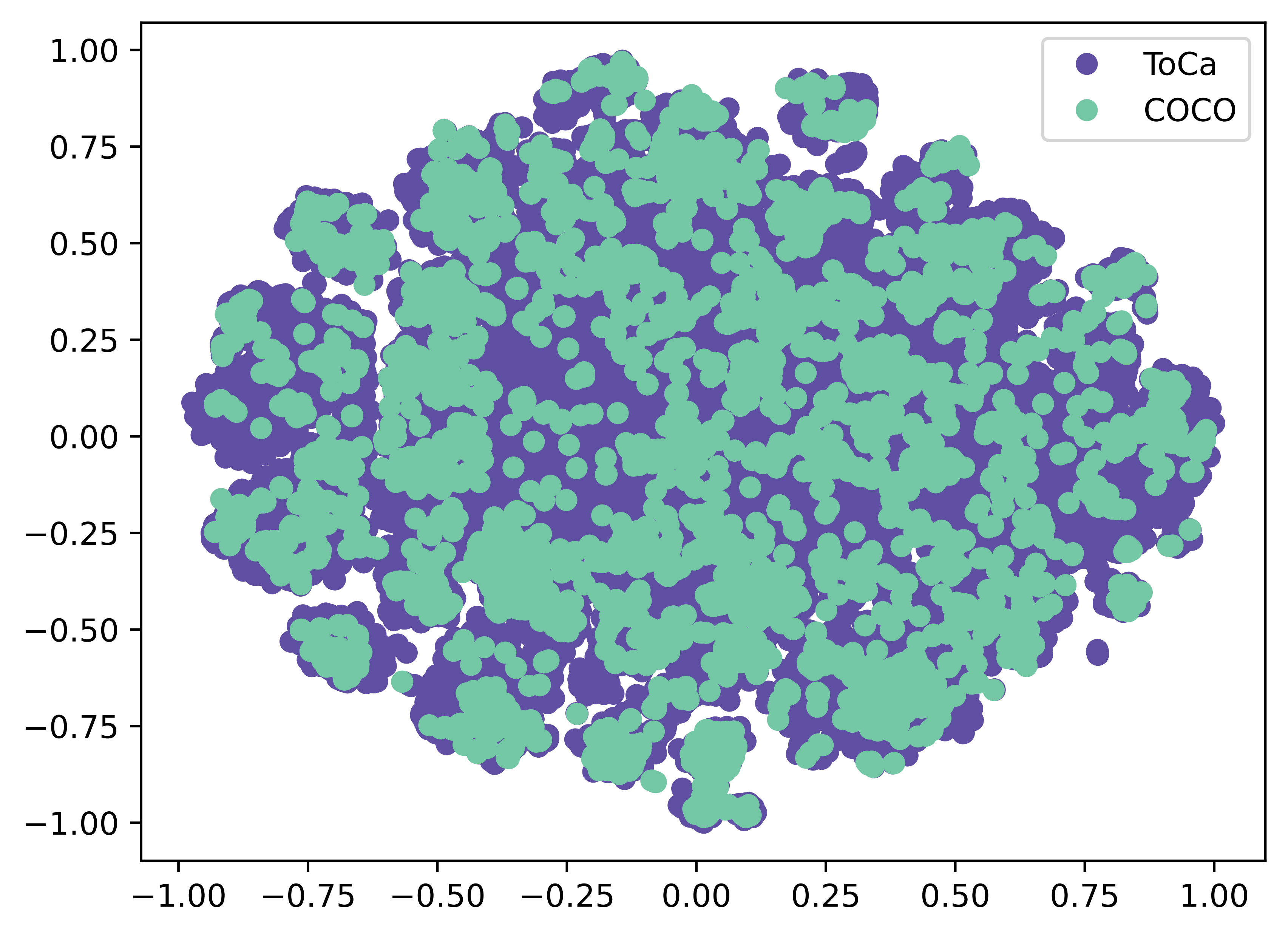}
        \label{fig:coco-toca-10x}
        }
      \end{subfigure}

    \vspace{0pt}
    \caption{t-SNE visualizations of Flickr30k, COCO, and ToCa. (a)-(c) and (d)-(f) respectively represent the variations in the distribution relations of features between Flickr30k and COCO, as well as between ToCa and COCO, across a relative quantity range from $0.1\times$ to $1\times$ to $10\times$.}
    \label{fig:tsne}
  \end{figure}
  
To visually showcase the distance between the synthesized data by ToCa and the distribution of the target data, t-SNE~\cite{tsne} is employed to visualize the text features encoded by CLIP in $\mathcal{D}$ and $\mathcal{T}$ as shown in Figure~\ref{fig:tsne}. Taking into consideration computational efficiency, a random selection of 5,667 text features from COCO is utilized as the analytical sample. The distribution relations between the ToCa-synthesized data and COCO data are visualized across a relative quantity range from $0.1\times$ to $1\times$ to $10\times$, as depicted in Figure~\ref{fig:tsne} (d)-(f). For ease of observation, the distributional relationship of data features between the two datasets, Flickr30k and COCO, used in the cross-domain captioning setting, is also visualized and presented as a reference.
Upon comparing Figure~\ref{fig:tsne} (a)-(c) and (d)-(f), it becomes evident that the degree of overlap between the synthesized data of ToCa and the distribution of COCO data is significantly higher than the overlap between the distribution of Flickr30k and COCO data. Simultaneously, upon observing the transition from (d) to (f), with an increase in data volume, ToCa exhibits an increasing number of novel samples. However, the proximity of these novel samples to COCO surpasses that of Flickr30k and COCO. Consequently, this outcome can qualitatively affirm that ToCa represents an effective solution for solving the objective Eq~\ref{eq:objective}, wherein it not only approximates the target data but also generates novel samples surpassing the target data.

In addition to the qualitative analysis of text feature visualization and the quantitative results regarding the final captioning performance, we have also conducted statistical measurements, as shown in Table~\ref{tab:stat}, as a direct evaluation of the quantitative indicators for assessing the distance between the generated data and the target dataset. To facilitate analysis, we have also included the statistical measurements of Flickr30k on COCO as a reference for comparison.
The statistical measurement metrics include precision (P) and recall (R) for both tokens and structures. Additionally, weighted precision ($\text{P}_w$) and weighted recall ($\text{R}_w$) are computed, taking into account the frequency. Furthermore, the cosine similarity ($\text{Cosine}$) is calculated for the frequency distribution of tokens and structures between the two datasets.
From Table~\ref{tab:stat}, it can be observed that the ($\text{P}_w$) and ($\text{R}_w$) metrics, which take frequency into consideration, are significantly higher than the $\text{P}$ and $\text{R}$ metrics.
This indicates that COCO dataset consists of both high-frequency and low-frequency data, and ToCa effectively adheres to these characteristics.
Comparing the results of ToCa and Flickr30k, it can be observed that the level of approximation between the synthesized data by ToCa and COCO is higher than that of Flickr30k and COCO (as indicated by higher $\text{R}_w$ and Cosine values for ToCa). Additionally, it is evident that the token-based Cosine for Flickr30k are lower than the structure-based. This is because Flickr30k and COCO involve descriptions of different domains of images, resulting in significant token differences. The higher structure-based Cosine can be attributed to the fact that both datasets involve captioning tasks, where the text follows a caption-style format.
Both token and structure metrics for the data generated by ToCa exhibit high values ($\text{R}_w$ and Cosine). Furthermore, it can be observed that the P values for both tokens and structures are significantly lower for ToCa compared to Flickr30k. This is primarily due to the larger volume of data generated by ToCa, resulting in the generation of more novel tokens and structures.

\begin{table}

    \begin{center}
    \caption{Statistical measures between $\mathcal{D}$ and $\mathcal{T}$.}
    \vspace{10pt}

    \begin{tabular}{l|ccccc|ccccc}
    \toprule
        ~ & \multicolumn{5}{c|}{ToCa on COCO} & \multicolumn{5}{c}{Flickr30k on COCO} \\
        ~ & P & R & $\text{P}_w$ & $\text{R}_w$ & Cosine
            & P & R & $\text{P}_w$ & $\text{R}_w$ & Cosine  \\
    \midrule
        token 
            & 43.8 & 59.0 & 99.4 & 99.6 & 92.7 
                & 73.9 & 23.5 & 98.8 & 47.2 & 10.6 \\
        structure 
            & 3.9 & 48.1 & 66.3 & 93.5 & 94.2 
                & 26.7 & 19.0 & 81.6 & 87.3 & 91.3 \\
    \bottomrule
    \end{tabular}
    \label{tab:stat}
    \end{center}

\end{table}

\section{Synthesized text examples}
\label{sec:examples}
We provide more examples of ToCa generated text in Table~\ref{tab:coco-examples}-\ref{tab:msrvtt-examples}, including the corresponding entity $token$, structure template $G_s$, sentence template $s'$, and generated text $s''$.
Tables~\ref{tab:coco-examples}, \ref{tab:flickr30k-examples}, \ref{tab:flickrstyle_h-examples}, \ref{tab:flickrstyle_r-examples} and \ref{tab:msrvtt-examples} show texts synthesized based on COCO, Flickr30k, FlickrStyle10k humor, FlickrStyle10k romantic, MSR-VTT as corpora, and texts from corpora, respectively. It can be seen that ToCa can synthesize different styles of text according to different corpora, and has strong flexibility and adaptability. The synthesized text also basically follows the token and template structure contained in prompt.

\newcolumntype{C}[1]{>{\centering\arraybackslash}m{#1}}

\begin{table}[hb]
    \centering
    \caption{The synthesized text examples with corpus $\mathcal{S}$ = COCO}
    \label{tab:coco-examples}
    \vspace{10pt}
    \begin{tabular}{C{0.1\linewidth}|m{0.9\linewidth}}
        \toprule
        Symbol & \multicolumn{1}{c}{Text} \\
        \midrule

        \textcolor{gray}{${token}$} & \textcolor{gray}{\text{['dining', 'area', 'chairs']}} \\
        \textcolor{gray}{${G_s}$} & \textcolor{gray}{\text{[J] [N] [VBN] with [J] [N] .}} \\
        \textcolor{gray}{${s'}$} & \textcolor{gray}{\text{[ ] dining [ ] area [ ] with [ ] chairs [ ] .}} \\
        $s''$ & The dining area is furnished with six chairs around a large table. \\
        \midrule

        \textcolor{gray}{${token}$} & \textcolor{gray}{\text{['elephant', 'walking', 'water']}} \\
        \textcolor{gray}{${G_s}$} & \textcolor{gray}{\text{[J] [N] with [N] [VBD] on .}} \\
        \textcolor{gray}{${s'}$} & \textcolor{gray}{\text{[ ] elephant [ ] walking [ ] with [ ] water [ ] on [ ] .}} \\
        $s''$ & An elephant walking through the jungle with water splashing on its back. \\
        \midrule

        \textcolor{gray}{${token}$} & \textcolor{gray}{\text{['different', 'fruits', 'sitting', 'table']}} \\
        \textcolor{gray}{${G_s}$} & \textcolor{gray}{\text{[J] [N] [VBG] on [N] .}} \\
        \textcolor{gray}{${s'}$} & \textcolor{gray}{\text{[ ] different [ ] fruits [ ] sitting [ ] on [ ] table [ ] .}} \\
        $s''$ & Three different fruits sitting on the kitchen table. \\
        \midrule

        \textcolor{gray}{${token}$} & \textcolor{gray}{\text{['close', 'plate', 'broccoli', 'table', 'food']}} \\
        \textcolor{gray}{${G_s}$} & \textcolor{gray}{\text{[J] [N] of [N] on [N] with [N] .}} \\
        \textcolor{gray}{${s'}$} & \textcolor{gray}{\text{[ ] close [ ] plate [ ] of [ ] broccoli [ ] on [ ] table [ ] with [ ] food [ ] .}} \\
        $s''$ & A close plate of broccoli sits on the table with other food. \\
        \midrule

        \textcolor{gray}{${token}$} & \textcolor{gray}{\text{['son', 'kite', 'beach']}} \\
        \textcolor{gray}{${G_s}$} & \textcolor{gray}{\text{[N] [J] in [N] off of [N] .}} \\
        \textcolor{gray}{${s'}$} & \textcolor{gray}{\text{[ ] son [ ] in [ ] kite [ ] off [ ] of [ ] beach [ ] .}} \\
        $s''$ & The son launches a kite from the edge of the sandy beach. \\
        \midrule

        \textcolor{gray}{${token}$} & \textcolor{gray}{\text{['woman', 'holding', 'white', 'tennis']}} \\
        \textcolor{gray}{${G_s}$} & \textcolor{gray}{\text{[N] [VBG] [J] [N] .}} \\
        \textcolor{gray}{${s'}$} & \textcolor{gray}{\text{[ ] woman [ ] holding [ ] white [ ] tennis [ ] .}} \\
        $s''$ & A woman holding a white tennis racket. \\
        \midrule

        \textcolor{gray}{${token}$} & \textcolor{gray}{\text{['man', 'front', 'sitting', 'table']}} \\
        \textcolor{gray}{${G_s}$} & \textcolor{gray}{\text{[N] of [N] [VBG] [N] .}} \\
        \textcolor{gray}{${s'}$} & \textcolor{gray}{\text{[ ] man [ ] of [ ] front [ ] sitting [ ] table [ ] .}} \\
        $s''$ & A man sitting at the front table of a caf. \\
        \midrule

        \textcolor{gray}{${token}$} & \textcolor{gray}{\text{['limit', 'sign']}} \\
        \textcolor{gray}{${G_s}$} & \textcolor{gray}{\text{[N] [VBG] [N] [VBN] in [N] of [N] .}} \\
        \textcolor{gray}{${s'}$} & \textcolor{gray}{\text{[ ] limit [ ] sign [ ] in [ ] of [ ] .}} \\
        $s''$ & The speed limit sign is in the center of the roundabout, limiting the vehicles' speed. \\
        \midrule

        \multirow{5}{*}{COCO} & A person holding a kite surrounded by a group of people. \\
            & A close up of a desert plate with a man sitting at the table. \\
            & A large truck parked on a city street. \\
            & Two teams playing football on green field with cars behind. \\
            & A group of young men playing a game of soccer. \\

        \bottomrule
    \end{tabular}
\end{table}

\newpage

\newcolumntype{C}[1]{>{\centering\arraybackslash}m{#1}}

\begin{table}
    \centering
    \caption{The synthesized text examples with corpus $\mathcal{S}$ = Flickr30k}
    \label{tab:flickr30k-examples}
    \vspace{10pt}
    \begin{tabular}{C{0.1\linewidth}|m{0.9\linewidth}}
        \toprule
        Symbol & \multicolumn{1}{c}{Text} \\
        \midrule

        \textcolor{gray}{${token}$} & \textcolor{gray}{\text{['sunglasses', 'building']}} \\
        \textcolor{gray}{${G_s}$} & \textcolor{gray}{\text{[N] on [N] .}} \\
        \textcolor{gray}{${s'}$} & \textcolor{gray}{\text{[ ] sunglasses [ ] on [ ] building [ ] .}} \\
        $s''$ & Sunglasses placed on the window of a modern building. \\
        \midrule

        \textcolor{gray}{${token}$} & \textcolor{gray}{\text{['seven', 'water', 'people', 'table']}} \\
        \textcolor{gray}{${G_s}$} & \textcolor{gray}{\text{[R] [VBN] [N] [VBZ] [N] with [N] .}} \\
        \textcolor{gray}{${s'}$} & \textcolor{gray}{\text{[ ] seven [ ] water [ ] people [ ] with [ ] table [ ] .}} \\
        $s''$ & Seven people enjoying water at a table for a picnic. \\
        \midrule

        \textcolor{gray}{${token}$} & \textcolor{gray}{\text{['woman', 'pier', 'water']}} \\
        \textcolor{gray}{${G_s}$} & \textcolor{gray}{\text{[N] with [N] [VBZ] in [N] .}} \\
        \textcolor{gray}{${s'}$} & \textcolor{gray}{\text{[ ] woman [ ] with [ ] pier [ ] in [ ] water [ ] .}} \\
        $s''$ & A woman fishing with a pierced lip in the clear waters of a lake. \\
        \midrule

        \textcolor{gray}{${token}$} & \textcolor{gray}{\text{['older', 'shirt', 'standing', 'front']}} \\
        \textcolor{gray}{${G_s}$} & \textcolor{gray}{\text{[J] [N] [VBG] in [N] .}} \\
        \textcolor{gray}{${s'}$} & \textcolor{gray}{\text{[ ] older [ ] shirt [ ] standing [ ] in [ ] front [ ] .}} \\
        $s''$ & An older man wearing a faded blue shirt is standing in front of the crowd. \\
        \midrule

        \textcolor{gray}{${token}$} & \textcolor{gray}{\text{['teams', 'playing', 'field']}} \\
        \textcolor{gray}{${G_s}$} & \textcolor{gray}{\text{[N] [VBG] [N] by .}} \\
        \textcolor{gray}{${s'}$} & \textcolor{gray}{\text{[ ] teams [ ] playing [ ] field [ ] by [ ] .}} \\
        $s''$ & Two baseball teams playing on a sunny field by the umpire. \\
        \midrule

        \textcolor{gray}{${token}$} & \textcolor{gray}{\text{['young', 'girl', 'boy', 'shirt']}} \\
        \textcolor{gray}{${G_s}$} & \textcolor{gray}{\text{[J] [N] in [N] of [N] .}} \\
        \textcolor{gray}{${s'}$} & \textcolor{gray}{\text{[ ] young [ ] girl [ ] in [ ] boy [ ] of [ ] shirt [ ] .}} \\
        $s''$ & A young girl in a red shirt is playing with a boy wearing a blue shirt. \\
        \midrule

        \textcolor{gray}{${token}$} & \textcolor{gray}{\text{['child', 'playing', 'grass', 'wearing', 'front']}} \\
        \textcolor{gray}{${G_s}$} & \textcolor{gray}{\text{[N] [VBG] [N] while [VBG] [N] .}} \\
        \textcolor{gray}{${s'}$} & \textcolor{gray}{\text{[ ] child [ ] playing [ ] grass [ ] while [ ] wearing [ ] front [ ] .}} \\
        $s''$ & A child playing in the green grass while wearing a red front shirt. \\
        \midrule

        \textcolor{gray}{${token}$} & \textcolor{gray}{\text{['young', 'front', 'girl', 'shirt']}} \\
        \textcolor{gray}{${G_s}$} & \textcolor{gray}{\text{[J] [N] for [N] of [N] .}} \\
        \textcolor{gray}{${s'}$} & \textcolor{gray}{\text{[ ] young [ ] front [ ] for [ ] girl [ ] of [ ] shirt [ ] .}} \\
        $s''$ & A young girl stands in front, holding up a red shirt for her friend to see. \\
        \midrule

        \textcolor{gray}{${token}$} & \textcolor{gray}{\text{['pretty', 'woman', 'wearing', 'dress']}} \\
        \textcolor{gray}{${G_s}$} & \textcolor{gray}{\text{[J] [N] [VBG] over [N] .}} \\
        \textcolor{gray}{${s'}$} & \textcolor{gray}{\text{[ ] pretty [ ] woman [ ] wearing [ ] over [ ] dress [ ] .}} \\
        $s''$ & A pretty woman wearing a red overcoat over a floral dress is walking in the park. \\
        \midrule

        \textcolor{gray}{${token}$} & \textcolor{gray}{\text{['shorts', 'green', 'ball', 'hit', 'background', 'man', 'table']}} \\
        \textcolor{gray}{${G_s}$} & \textcolor{gray}{\text{[N] in [J] [N] [VB] [N] with [N] while [J] [N] [VBP] .}} \\
        \textcolor{gray}{${s'}$} & \textcolor{gray}{\text{[ ] shorts [ ] in [ ] green [ ] ball [ ] hit [ ] background [ ] with [ ] man [ ] while [ ] table [ ] .}} \\
        $s''$ & A man wearing shorts hits a green ball against the background with a table nearby. \\
        \midrule

        \textcolor{gray}{${token}$} & \textcolor{gray}{\text{['singing', 'white', 'man', 'playing', 'guitar']}} \\
        \textcolor{gray}{${G_s}$} & \textcolor{gray}{\text{[N] in [J] [N] [VBG] [N] .}} \\
        \textcolor{gray}{${s'}$} & \textcolor{gray}{\text{[ ] singing [ ] in [ ] white [ ] man [ ] playing [ ] guitar [ ] .}} \\
        $s''$ & A white man playing the guitar and singing in a quiet room. \\
        \midrule

        \multirow{3}{*}{Flickr30k} & Two black guys, that are on rival teams playing basketball. \\
            & A white man playing an acoustic guitar with another person. \\
            & A young girl in a yellow shirt is holding a bird for a speaker. \\

        \bottomrule
    \end{tabular}
\end{table}

\newpage

\newcolumntype{C}[1]{>{\centering\arraybackslash}m{#1}}

\begin{table}
    \centering
    \caption{The synthesized text examples with corpus $\mathcal{S}$ = FlickrStyle10k humor}
    \label{tab:flickrstyle_h-examples}
    \vspace{10pt}
    \begin{tabular}{C{0.1\linewidth}|m{0.9\linewidth}}
        \toprule
        Symbol & \multicolumn{1}{c}{Text} \\
        \midrule

        \textcolor{gray}{${token}$} & \textcolor{gray}{\text{['young', 'pink', 'shirt']}} \\
        \textcolor{gray}{${G_s}$} & \textcolor{gray}{\text{[J] [N] and [J] [N] [VBG] into [J] [N] .}} \\
        \textcolor{gray}{${s'}$} & \textcolor{gray}{\text{[ ] young [ ] pink [ ] and [ ] shirt [ ] into [ ] .}} \\
        $s''$ & A young girl in a pink dress and a white shirt runs into the garden. \\
        \midrule        
        
        \textcolor{gray}{${token}$} & \textcolor{gray}{\text{['man', 'walk', 'street']}} \\
        \textcolor{gray}{${G_s}$} & \textcolor{gray}{\text{[N] [VBP] [R] [J] [N] after [VBN] by [N] .}} \\
        \textcolor{gray}{${s'}$} & \textcolor{gray}{\text{[ ] man [ ] walk [ ] street [ ] after [ ] by [ ] .}} \\
        $s''$ & A man walks down the street after sunset. \\
        \midrule

        \textcolor{gray}{${token}$} & \textcolor{gray}{\text{['red', 'ball', 'water', 'trying']}} \\
        \textcolor{gray}{${G_s}$} & \textcolor{gray}{\text{[J] [N] at [N] [VBG] like [VBZ] [J] .}} \\
        \textcolor{gray}{${s'}$} & \textcolor{gray}{\text{[ ] red [ ] ball [ ] at [ ] water [ ] trying [ ] like [ ] .}} \\
        $s''$ & A red ball floating at the surface of the water, trying to stay afloat. \\
        \midrule

        \textcolor{gray}{${token}$} & \textcolor{gray}{\text{['dog', 'bones']}} \\
        \textcolor{gray}{${G_s}$} & \textcolor{gray}{\text{[N] with [N] [VBZ] in [N] , [VBG] there [N] .}} \\
        \textcolor{gray}{${s'}$} & \textcolor{gray}{\text{[ ] dog [ ] with [ ] bones [ ] in [ ] , [ ] there [ ] .}} \\
        $s''$ & A dog chewing on bones, scattered around its den, in the forest. \\
        \midrule

        \textcolor{gray}{${token}$} & \textcolor{gray}{\text{['child', 'blue', 'shirt', 'white']}} \\
        \textcolor{gray}{${G_s}$} & \textcolor{gray}{\text{[N] under [J] [N] [VB] [J] for [VBG] .}} \\
        \textcolor{gray}{${s'}$} & \textcolor{gray}{\text{[ ] child [ ] under [ ] blue [ ] shirt [ ] white [ ] for [ ] .}} \\
        $s''$ & A child wearing a blue shirt and a white sweater is playing for forty-five minutes. \\
        \midrule

        \textcolor{gray}{${token}$} & \textcolor{gray}{\text{['jacket', 'playing']}} \\
        \textcolor{gray}{${G_s}$} & \textcolor{gray}{\text{[N] of [N] [VBG] [R] [J] and [J] [N] [VBG] for [N] .}} \\
        \textcolor{gray}{${s'}$} & \textcolor{gray}{\text{[ ] jacket [ ] of [ ] playing [ ] and [ ] for [ ] .}} \\
        $s''$ & A red jacket being worn by two children playing tag in the park during recess. \\
        \midrule

        \textcolor{gray}{${token}$} & \textcolor{gray}{\text{['white', 'dog', 'running', 'grass', 'looking', 'bones']}} \\
        \textcolor{gray}{${G_s}$} & \textcolor{gray}{\text{[J] [N] [VBG] above [N] [VBG] for [N] .}} \\
        \textcolor{gray}{${s'}$} & \textcolor{gray}{\text{[ ] white [ ] dog [ ] running [ ] above [ ] grass [ ] looking [ ] for [ ] bones [ ] .}} \\
        $s''$ & A white dog running above the green grass, looking for buried bones. \\
        \midrule

        \textcolor{gray}{${token}$} & \textcolor{gray}{\text{['man', 'looking', 'pokemon', 'go']}} \\
        \textcolor{gray}{${G_s}$} & \textcolor{gray}{\text{[N] [VBG] [N] in [N] [VBD] [N] .}} \\
        \textcolor{gray}{${s'}$} & \textcolor{gray}{\text{[ ] man [ ] looking [ ] pokemon [ ] in [ ] go [ ] .}} \\
        $s''$ & A man looking for a Pokemon Go game character in a park. \\
        \midrule

        \textcolor{gray}{${token}$} & \textcolor{gray}{\text{['man', 'other', 'people', 'grass']}} \\
        \textcolor{gray}{${G_s}$} & \textcolor{gray}{\text{[N] in [J] [N] [VBG] on [N] in from of [N] [VBD] [N] [J] .}} \\
        \textcolor{gray}{${s'}$} & \textcolor{gray}{\text{[ ] man [ ] in [ ] other [ ] people [ ] on [ ] grass [ ] in [ ] from [ ] of [ ] .}} \\
        $s''$ & A man watching other people playing football on the grass from the sidelines. \\
        \midrule

        \textcolor{gray}{${token}$} & \textcolor{gray}{\text{['water', 'fish']}} \\
        \textcolor{gray}{${G_s}$} & \textcolor{gray}{\text{[N] and [J] [N] on [N] [VBG] for [N] [VB] .}} \\
        \textcolor{gray}{${s'}$} & \textcolor{gray}{\text{[ ] water [ ] and [ ] fish [ ] on [ ] for [ ] .}} \\
        $s''$ & A large aquarium filled with colorful fish and clear water on the living room table. \\
        \midrule

        \textcolor{gray}{${token}$} & \textcolor{gray}{\text{['dog', 'ball', 'trying', 'win']}} \\
        \textcolor{gray}{${G_s}$} & \textcolor{gray}{\text{[N] over [N] of [J] [VBG] [VB] [J] [N] .}} \\
        \textcolor{gray}{${s'}$} & \textcolor{gray}{\text{[ ] dog [ ] over [ ] ball [ ] of [ ] trying [ ] win [ ] .}} \\
        $s''$ & A dog eagerly chasing after a red ball, trying to win it over. \\
        \midrule

        \multirow{3}{*}{humor} & A young girl in a pink dress with something squishy in her hand squeals. \\
            & Two men are approaching a gold colored statue of gandhi looking for pokemon go. \\
            & A child in a green shirt and crocks looking down a drain for monsters. \\

        \bottomrule
    \end{tabular}
\end{table}

\newpage

\newcolumntype{C}[1]{>{\centering\arraybackslash}m{#1}}

\begin{table}
    \centering
    \caption{The synthesized text examples with corpus $\mathcal{S}$ = FlickrStyle10k romantic}
    \label{tab:flickrstyle_r-examples}
    \vspace{10pt}
    \begin{tabular}{C{0.1\linewidth}|m{0.9\linewidth}}
        \toprule
        Symbol & \multicolumn{1}{c}{Text} \\
        \midrule

        \textcolor{gray}{${token}$} & \textcolor{gray}{\text{['surfer', 'riding', 'wave']}} \\
        \textcolor{gray}{${G_s}$} & \textcolor{gray}{\text{[N] [VBG] with [J] [N] that [VBG] .}} \\
        \textcolor{gray}{${s'}$} & \textcolor{gray}{\text{[ ] surfer [ ] riding [ ] with [ ] wave [ ] that [ ] .}} \\
        $s''$ & A surfer riding the crest of a towering wave, its white foamy tips crashing around him. \\
        \midrule

        \textcolor{gray}{${token}$} & \textcolor{gray}{\text{['huge', 'wave']}} \\
        \textcolor{gray}{${G_s}$} & \textcolor{gray}{\text{[J] [N] for [N] [VBN] [R] by [N] on [N] .}} \\
        \textcolor{gray}{${s'}$} & \textcolor{gray}{\text{[ ] huge [ ] wave [ ] for [ ] by [ ] on [ ] .}} \\
        $s''$ & A huge wave crashing against the shore for hours, relentlessly pounded by the strong winds. \\
        \midrule

        \textcolor{gray}{${token}$} & \textcolor{gray}{\text{['back', 'owner']}} \\
        \textcolor{gray}{${G_s}$} & \textcolor{gray}{\text{[R] [VBG] [N] [VBZ] [VBN] near [N] .}} \\
        \textcolor{gray}{${s'}$} & \textcolor{gray}{\text{[ ] back [ ] owner [ ] near [ ] .}} \\
        $s''$ & The lost item is close to the owner's back. \\
        \midrule

        \textcolor{gray}{${token}$} & \textcolor{gray}{\text{['man', 'holding', 'camera', 'love']}} \\
        \textcolor{gray}{${G_s}$} & \textcolor{gray}{\text{[N] [VBG] [VBN] in [N] , [VBG] for [N] .}} \\
        \textcolor{gray}{${s'}$} & \textcolor{gray}{\text{[ ] man [ ] holding [ ] in [ ] camera [ ] , [ ] for [ ] love [ ] .}} \\
        $s''$ & A man holding a camera in front of a beautiful sunset, capturing the moment for love. \\
        \midrule

        \textcolor{gray}{${token}$} & \textcolor{gray}{\text{['couple', 'love', 'street']}} \\
        \textcolor{gray}{${G_s}$} & \textcolor{gray}{\text{[N] and [J] [N] [VBG] [N] for [N] .}} \\
        \textcolor{gray}{${s'}$} & \textcolor{gray}{\text{[ ] couple [ ] and [ ] love [ ] street [ ] for [ ] .}} \\
        $s''$ & A couple falls in love strolling down the vibrant and colorful Flower Street every evening. \\
        \midrule

        \textcolor{gray}{${token}$} & \textcolor{gray}{\text{['stand', 'waiting']}} \\
        \textcolor{gray}{${G_s}$} & \textcolor{gray}{\text{[VBP] [VBG] for [N] [VBG] on [N] and [VBG] [N] .}} \\
        \textcolor{gray}{${s'}$} & \textcolor{gray}{\text{[ ] stand [ ] waiting [ ] for [ ] on [ ] and [ ] .}} \\
        $s''$ & Three people stand at the bus stop, waiting for the bus to arrive on a rainy corner. \\
        \midrule

        \textcolor{gray}{${token}$} & \textcolor{gray}{\text{['play', 'snow']}} \\
        \textcolor{gray}{${G_s}$} & \textcolor{gray}{\text{[N] and [N] [VBG] on [N] [VBG] [N] neither [VBG] how [VB] .}} \\
        \textcolor{gray}{${s'}$} & \textcolor{gray}{\text{[ ] play [ ] and [ ] snow [ ] on [ ] neither [ ] how [ ] .}} \\
        $s''$ & People play in the snow neither melts nor sticks to the ground. \\
        \midrule

        \textcolor{gray}{${token}$} & \textcolor{gray}{\text{['bicycle', 'ramp']}} \\
        \textcolor{gray}{${G_s}$} & \textcolor{gray}{\text{[N] in [J] [N] [VBP] at [N] , [VBG] [J] [VBZ] .}} \\
        \textcolor{gray}{${s'}$} & \textcolor{gray}{\text{[ ] bicycle [ ] in [ ] ramp [ ] at [ ] , [ ] .}} \\
        $s''$ & A bicycle rolls down a steep ramp at the park. \\
        \midrule

        \textcolor{gray}{${token}$} & \textcolor{gray}{\text{['girl', 'stands', 'next', 'water']}} \\
        \textcolor{gray}{${G_s}$} & \textcolor{gray}{\text{[N] [VBZ] in [J] [N] and [VB] [N] below as [VBZ] [N] .}} \\
        \textcolor{gray}{${s'}$} & \textcolor{gray}{\text{[ ] girl [ ] stands [ ] in [ ] next [ ] water [ ] and [ ] below [ ] as [ ] .}} \\
        $s''$ & A girl stands next to the shore, gazes at the calm water and reflects deeply. \\
        \midrule

        \textcolor{gray}{${token}$} & \textcolor{gray}{\text{['white', 'hat']}} \\
        \textcolor{gray}{${G_s}$} & \textcolor{gray}{\text{[J] [N] , [VBG] [R] .}} \\
        \textcolor{gray}{${s'}$} & \textcolor{gray}{\text{[ ] white [ ] hat [ ] , [ ] .}} \\
        $s''$ & A man wearing a white sun hat, shielding his eyes from the bright sunlight. \\
        \midrule

        \textcolor{gray}{${token}$} & \textcolor{gray}{\text{['jumps', 'pool', 'water']}} \\
        \textcolor{gray}{${G_s}$} & \textcolor{gray}{\text{[N] [VBZ] [N] and [VBZ] with [N] [VB] [VBG] .}} \\
        \textcolor{gray}{${s'}$} & \textcolor{gray}{\text{[ ] jumps [ ] pool [ ] and [ ] with [ ] water [ ] .}} \\
        $s''$ & Someone jumps into a swimming pool and splashes around with water. \\
        \midrule

        \multirow{3}{*}{romantic} & A group of people surf on a large wave to experience the thrill of life. \\
            & A man skateboarding in an empty swimming pool practicing for a tournament. \\
            & The child is holding a camera up close to capture a scene of peace and love. \\
                      
        \bottomrule
    \end{tabular}
\end{table}

\newpage

\newcolumntype{C}[1]{>{\centering\arraybackslash}m{#1}}

\begin{table}
    \centering
    \caption{The synthesized text examples with corpus $\mathcal{S}$ = MSR-VTT}
    \label{tab:msrvtt-examples}
    \vspace{10pt}
    \begin{tabular}{C{0.1\linewidth}|m{0.9\linewidth}}
        \toprule
        Symbol & \multicolumn{1}{c}{Text} \\
        \midrule

        \textcolor{gray}{${token}$} & \textcolor{gray}{\text{['people', 'dance', 'stage']}} \\
        \textcolor{gray}{${G_s}$} & \textcolor{gray}{\text{[N] [VBP] around [N] and [VB] [J] [N] .}} \\
        \textcolor{gray}{${s'}$} & \textcolor{gray}{\text{[ ] people [ ] dance [ ] around [ ] stage [ ] and [ ] .}} \\
        $s''$ & Several people dance around an elaborate stage setup and sing into microphones. \\
        \midrule

        \textcolor{gray}{${token}$} & \textcolor{gray}{\text{['boy', 'hugging', 'woman']}} \\
        \textcolor{gray}{${G_s}$} & \textcolor{gray}{\text{[N] [VBG] on [N] .}} \\
        \textcolor{gray}{${s'}$} & \textcolor{gray}{\text{[ ] boy [ ] hugging [ ] on [ ] woman [ ] .}} \\
        $s''$ & A young boy hugging an elderly woman on the park bench. \\
        \midrule

        \textcolor{gray}{${token}$} & \textcolor{gray}{\text{['man', 'ted', 'talk']}} \\
        \textcolor{gray}{${G_s}$} & \textcolor{gray}{\text{[N] of [J] [N] [VBP] .}} \\
        \textcolor{gray}{${s'}$} & \textcolor{gray}{\text{[ ] man [ ] of [ ] ted [ ] talk [ ] .}} \\
        $s''$ & A man giving a thought-provoking TED talk on technology. \\
        \midrule

        \textcolor{gray}{${token}$} & \textcolor{gray}{\text{['sandwich', 'head']}} \\
        \textcolor{gray}{${G_s}$} & \textcolor{gray}{\text{[N] [VBZ] on [N] in [N] .}} \\
        \textcolor{gray}{${s'}$} & \textcolor{gray}{\text{[ ] sandwich [ ] on [ ] head [ ] in [ ] .}} \\
        $s''$ & A man accidentally drops a sandwich on another man's head at the picnic. \\
        \midrule

        \textcolor{gray}{${token}$} & \textcolor{gray}{\text{['tiger', 'killing', 'man']}} \\
        \textcolor{gray}{${G_s}$} & \textcolor{gray}{\text{[N] [VBG] [N] and [VB] about .}} \\
        \textcolor{gray}{${s'}$} & \textcolor{gray}{\text{[ ] tiger [ ] killing [ ] man [ ] and [ ] about [ ] .}} \\
        $s''$ & A tiger on the loose, killing a man and leaving authorities searching for it in the forest. \\
        \midrule

        \textcolor{gray}{${token}$} & \textcolor{gray}{\text{['kid', 'playing', 'game', 'other', 'video', 'car']}} \\
        \textcolor{gray}{${G_s}$} & \textcolor{gray}{\text{[N] [VBG] [N] [J] [N] at [N] .}} \\
        \textcolor{gray}{${s'}$} & \textcolor{gray}{\text{[ ] kid [ ] playing [ ] game [ ] other [ ] video [ ] at [ ] car [ ] .}} \\
        $s''$ & A kid playing a video game on a car seat next to another kid. \\
        \midrule

        \textcolor{gray}{${token}$} & \textcolor{gray}{\text{['vine', 'compilation']}} \\
        \textcolor{gray}{${G_s}$} & \textcolor{gray}{\text{[J] [N] and and [VBG] .}} \\
        \textcolor{gray}{${s'}$} & \textcolor{gray}{\text{[ ] vine [ ] compilation [ ] and [ ] and [ ] .}} \\
        $s''$ & A vine compilation of honeysuckle and jasmine flowers growing entwined around a trellis. \\
        \midrule

        \textcolor{gray}{${token}$} & \textcolor{gray}{\text{['person', 'discussing', 's', 'game']}} \\
        \textcolor{gray}{${G_s}$} & \textcolor{gray}{\text{[N] [VBG] [N] and [N] .}} \\
        \textcolor{gray}{${s'}$} & \textcolor{gray}{\text{[ ] person [ ] discussing [ ] s [ ] and [ ] game [ ] .}} \\
        $s''$ & A person is passionately discussing the strategies and rules of a chess game. \\
        \midrule

        \textcolor{gray}{${token}$} & \textcolor{gray}{\text{['children', 'cartoon']}} \\
        \textcolor{gray}{${G_s}$} & \textcolor{gray}{\text{[N] [VBZ] [N] and [VBZ] [VBD] .}} \\
        \textcolor{gray}{${s'}$} & \textcolor{gray}{\text{[ ] children [ ] cartoon [ ] and [ ] .}} \\
        $s''$ & Several children watching an animated cartoon show together. \\
        \midrule

        \textcolor{gray}{${token}$} & \textcolor{gray}{\text{['feet', 'water']}} \\
        \textcolor{gray}{${G_s}$} & \textcolor{gray}{\text{[N] from [J] [N] .}} \\
        \textcolor{gray}{${s'}$} & \textcolor{gray}{\text{[ ] feet [ ] from [ ] water [ ] .}} \\
        $s''$ & The feet of the swimmer are only a few inches from the crystal-clear water. \\
        \midrule

        \textcolor{gray}{${token}$} & \textcolor{gray}{\text{['animated', 'cartoon', 'man']}} \\
        \textcolor{gray}{${G_s}$} & \textcolor{gray}{\text{[J] [N] about [N] .}} \\
        \textcolor{gray}{${s'}$} & \textcolor{gray}{\text{[ ] animated [ ] cartoon [ ] about [ ] man [ ] .}} \\
        $s''$ & An animated cartoon about a man who becomes a superhero. \\
        \midrule

        \multirow{3}{*}{MSR-VTT} & People are playing a chess-like video game. \\
            & An animated cartoon about a guy looking for dinosaurs. \\
            & A lion is attacking a hippo and then a man talks about a national park. \\

        \bottomrule
    \end{tabular}
\end{table}

\clearpage

\end{document}